\documentclass[11pt]{article}

\usepackage[final]{acl}

\usepackage{times}
\usepackage{latexsym}

\usepackage[T1]{fontenc}
\usepackage[utf8]{inputenc}

\usepackage{microtype}

\usepackage{inconsolata}

\usepackage{graphicx}

\usepackage{multirow}
\usepackage{tabularx}
\usepackage{wrapfig}
\usepackage{placeins}
\usepackage{colortbl}
\usepackage{pifont}
\usepackage{makecell}
\newcommand{\cmark}{\ding{51}}
\newcommand{\xmark}{\ding{55}}
\usepackage{amssymb}
\usepackage{amsfonts}
\usepackage{xurl}
\usepackage{booktabs}
\usepackage{nicefrac}
\usepackage{xcolor}
\usepackage{enumitem}
\usepackage{tikz}
\usepackage{pgfplots}
\usepackage{xspace}
\usepackage[most]{tcolorbox}
\usepackage{etoc}

\newcommand{\ours}{\textsc{InfoEdit}\xspace}

\title{\ours: Probing Global Layout Reasoning in Infographic Editing}

\author{
  Cheng Yang\textsuperscript{1,}\thanks{\hspace{1mm}Equal contribution. Project Page:~\url{https://infoedit.github.io}} \quad
  Chufan Shi\textsuperscript{2,*} \quad
  Huijuan Wang\textsuperscript{2,*} \quad
  Bo Shui\textsuperscript{3} \\
  \textbf{Yaokang Wu\textsuperscript{4}} \quad
  \textbf{Muzi Tao\textsuperscript{2}} \quad
  \textbf{Yibo Yan\textsuperscript{2}} \quad
  \textbf{Xuezhe Ma\textsuperscript{2}} \quad
  \textbf{Taylor Berg-Kirkpatrick\textsuperscript{1}} \\
  \textsuperscript{1}University of California San Diego \quad
  \textsuperscript{2}University of Southern California \\
  \textsuperscript{3}University of Illinois Urbana-Champaign \quad
  \textsuperscript{4}Carnegie Mellon University \\
  \texttt{chy085@ucsd.edu} \quad
  \texttt{\{chufansh,huijuanw\}@usc.edu}
}

\begin{document}
\maketitle

\etocdepthtag.toc{mainmatter}

\begin{abstract}
Multimodal foundation models edit natural photographs at production quality, yet the same models struggle with structured visual content such as infographics. Unlike photographs, infographics encode information through logical relations; editing one element often requires surrounding elements to be adapted. We refer to this global layout reasoning capability as \emph{reflow}. Existing image-editing benchmarks neither provide a dedicated setting for structured visual content nor evaluate the reflow capability.
We introduce \ours, a novel benchmark of $1{,}000$ infographics across eight logical-relation families, paired with $4{,}000$ editing instructions across four editing tasks, and a reflow-aware evaluation protocol. Across eight frontier editors, only GPT-Image-2 clears $60\%$ average success rate; most models fall below $7\%$, and no editor exceeds $36\%$ on the \textsc{Swap-Block} task even with perfect target localization. We further show that code-level editing can match the strongest pixel-level editor, revealing complementary strengths across tasks. \ours identifies reflow as a central challenge in structured visual content editing and provides a diagnostic benchmark to facilitate future progress.
\end{abstract}

\begin{figure*}[!t]
    \centering
    \includegraphics[width=\textwidth]{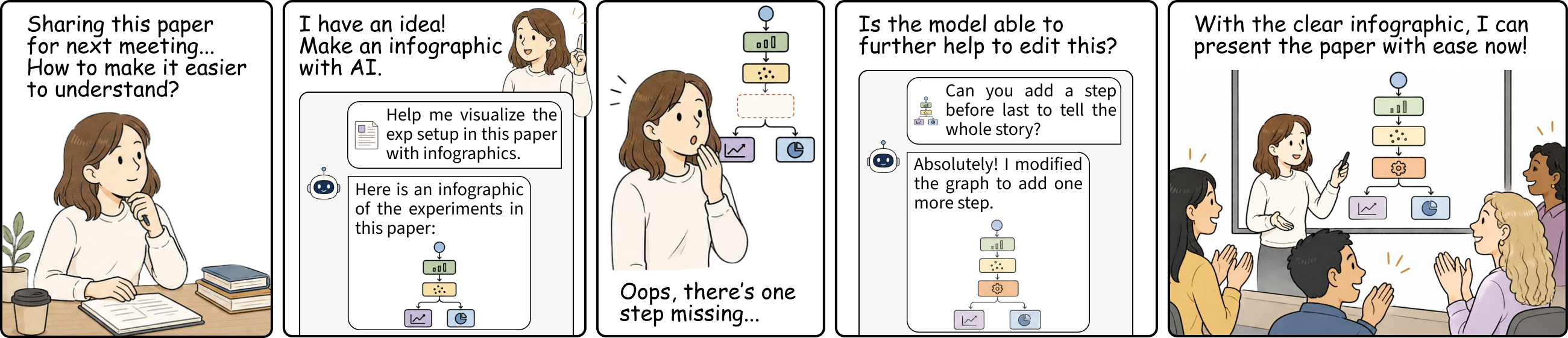} 
    \vspace{-1.5em}
    \caption{A typical ``vibe-design'' workflow with structured visual content. A user asks an image editor to add a missing step to an infographic. The instruction names one element, but the surrounding nodes and connectors must all reflow for the result to be usable. \textsc{InfoEdit} measures whether current editors can perform this reflow reliably.}
    \vspace{-1.5em}
    \label{fig:scenario}
\end{figure*}

\section{Introduction}
Multimodal foundation models such as GPT-Image-2~\cite{openai2026gptimage2} and Nanobanana-Pro~\cite{google2025gemini3image} have pushed natural-image editing towards production quality. They can faithfully execute instructions such as \emph{``remove the tree''}, producing edits that are visually coherent and semantically aligned with user intent~\cite{huang2025diffusion}. Such instructions are often local in effect: they modify a specified target while keeping the surrounding scene consistent with the source.

Meanwhile, these same models are increasingly capable of generating structured visual content, such as infographics~\cite{google2025stitch,NotebookLM,zhu2026paperbanana,zhu2026autofigure}. Editing such content, however, remains underexplored. Fig.~\ref{fig:scenario} sketches a typical workflow: the user asks the model to insert a missing step into an infographic. The instruction names one element, but correct execution touches many: downstream nodes shift and connectors reroute. We call this \emph{reflow}, a form of global layout reasoning that requires the model to automatically adapt the mentioned element and other implicitly affected elements (e.g., re-position and re-size), so the resulting image remains balanced, legible and logically valid.
Existing image-editing benchmarks~\cite{ brooks2023instructpix2pix,ye2025imgedit,pan2025wiseedit,yu2025anyedit} focus on natural-image editing; they leave infographic editing underexplored and do not evaluate the reflow capability it requires.

Therefore, we present \ours (Sec.~\ref{sec:2}), a novel benchmark for editing infographics. Such infographics cover eight logical-relation families: List, Process, Cycle, Hierarchy, Relationship, Matrix, Pyramid and Picture. Each family imposes its own reflow contract, such as re-routing arrows in a Process, or reparenting in a Hierarchy. \ours contains $1{,}000$ human-curated infographics spanning these families on diverse real-world topics, each paired with editable source files in HTML ($800$) or PowerPoint ($200$).
For each infographic, we annotate four editing tasks at increasing perturbation scope, from text-level changes to canvas-level restructurings: \textsc{Expand-Text}, \textsc{Insert-Element}, \textsc{Swap-Block}, and \textsc{Reshape-Canvas}, yielding $4{,}000$ instruction–infographic pairs in total.
Together, these infographics and instructions make genuine reflow a prerequisite.

Furthermore, we develop a reflow-aware evaluation protocol (Sec.~\ref{sec:eval}) that grades each task separately, decomposing every verdict into two binary checks: edit compliance, the named change is executed on its target; and content preservation, every other element survives intact.  We report success only when both hold. The decomposition matters because failures on such structured visual content include not only wrong target edits but also correct edits that silently damage unmentioned elements.

We benchmark eight frontier proprietary and open-source models on \ours (Sec.~\ref{sec:experiments}). 
Current models remain far from solving infographic editing that requires reflow.
Only GPT-Image-2 clears $60\%$ average success rate; the next two models reach $38$--$42\%$, and the remaining five, including all three open-weight baselines, fall below $7\%$.
Our evaluation shows that reflow failures are task-specific, e.g., \textsc{Swap-Block} is bottlenecked by edit compliance,  capping at $36.5\%$ across all eight.
Open-weight models lag especially far behind, with $0\%$ success rate on
\textsc{Insert-Element}, \textsc{Swap-Block}, and \textsc{Reshape-Canvas}.

Further diagnostics reveal where the bottlenecks arise (Sec.~\ref{sec:discussion}). Code-level editing with editable source and visual grounding reaches $61.6\%$ average success rate, just matching GPT-Image-2 at $62.0\%$, but the two pathways show complementary strengths across tasks.
Interactive selection hints improve \textsc{Swap-Block} success by only $3.2$--$7.4$ points, with no model exceeding $36\%$ even under perfect target localization, indicating that the bottleneck is
the reflow operation itself rather than merely target localization.
Difficulty and error analyses show that reflow is multi-faceted, spanning correct placement, neighbor-aware adjustment, and global style and structure preservation.

To sum up, we envision \ours{} as a novel comprehensive benchmark suite for guiding research on infographic editing. By providing editable-source infographics, multi-scale editing tasks, and a reflow-aware evaluation framework, \ours{} not only evaluates how well foundation models execute the requested edit while preserving non-target content via reflow, but also offers diagnostic insights into their capabilities and failure modes when editing structured visual artifacts.

\section{\ours Benchmark}
\label{sec:2}
\subsection{Task Definition}
\label{sec:task}

\begin{figure*}[!t]
    \centering
    \includegraphics[width=\textwidth]{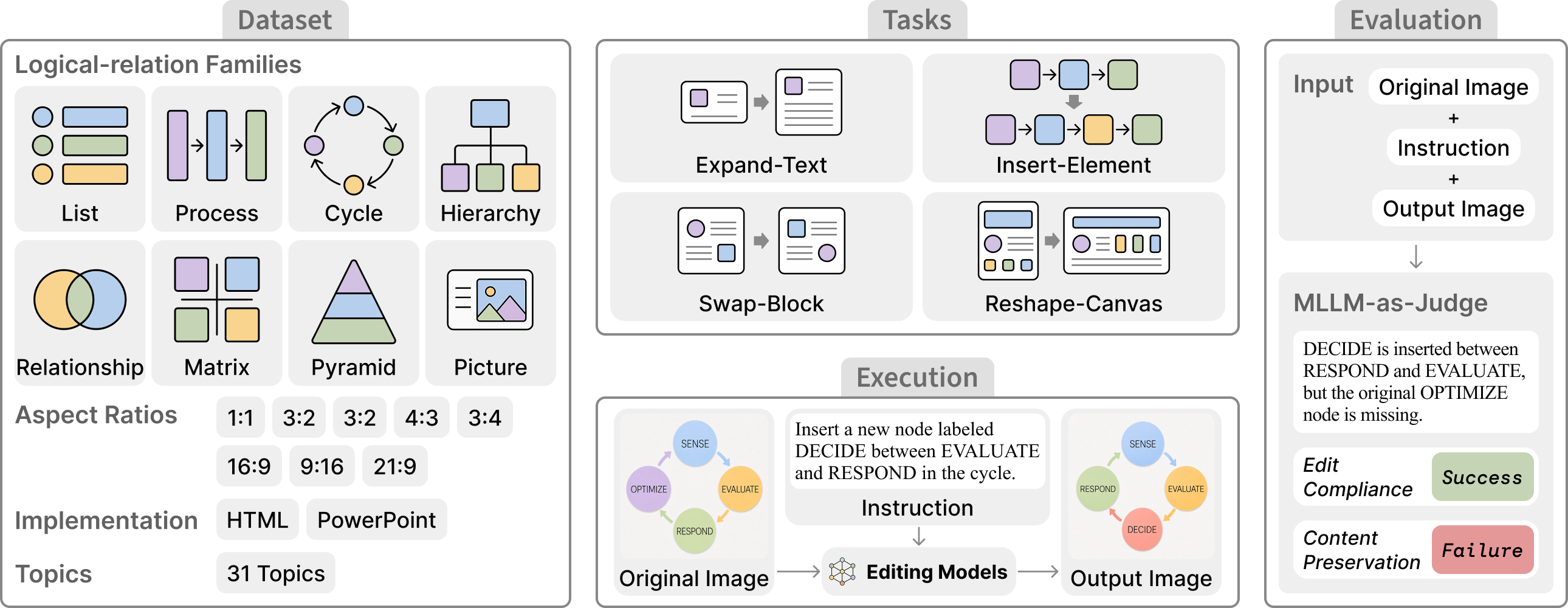} 
    \vspace{-1.5em}
\caption{Overview of \ours. We curate $1{,}000$ infographics covering eight logical-relation families (left), annotate four editing tasks for each infographic (middle), and evaluate model outputs on two binary aspects, edit compliance and content preservation (right). Data examples are provided in Appx.~\ref{appx:data_examples}.}
    \vspace{-1.5em}
    \label{fig:framework}
\end{figure*}

\ours assesses an image editor's ability to perform reflow editing on structured visual content. Given an original infographic $X$ and a natural-language editing instruction $I$ that names a structural change, the model $f$ must produce
\begin{equation*}
  X' = f(X, I) \quad \text{s.t.} \quad \mathcal{C}(X', X, I) \;\wedge\; \mathcal{P}(X', X, I).
\end{equation*}
$\mathcal{C}$ is the \emph{compliance} predicate: the change named by $I$ is faithfully executed in $X'$. $\mathcal{P}$ is the \emph{preservation} predicate: every element of $X$ not mentioned by $I$ remains present in $X'$, free to reflow to accommodate the change. We provide representative data examples of \ours in Appx.~\ref{appx:data_examples}.

\subsection{Infographics Construction}
\label{sec:data}

\textbf{Design Principles.}
Each infographic in \ours is designed around three principles that together support reliable evaluation. They are (i)~topically realistic, drawn from real-world infographic domains, so the benchmark reflects practical use cases. They are (ii)~aesthetically credible, with coherent palettes, balanced layouts, and typography on par with production work. Crucially, they are (iii)~logically structured: each infographic conveys information through visual logical-relation components.

\paragraph{Logical-Relation Taxonomy.}
Structured visual content conveys information through logical relations, which \ours organizes into eight families (Fig.~\ref{fig:framework}, left): List, Process, Cycle, Hierarchy, Relationship, Matrix, Pyramid, and Picture. These eight families cover common infographic structures used in everyday design tools~\cite{microsoft_smartart}; each encodes a distinct semantic relation and imposes reflow contracts that any valid edit must satisfy, such as reparenting in a Hierarchy.

\paragraph{Annotation Pipeline.}
\label{sec:pipeline}
The annotation pipeline runs in five stages and yields $1{,}000$ infographics, each implemented with editable, structurally addressable source files ($800$ HTML and $200$ PowerPoint). We summarize the pipeline below; full implementation details are provided in Appx.~\ref{appx:infographics_construction}.

\begin{figure*}[!t]
    \centering
    \includegraphics[width=\textwidth]{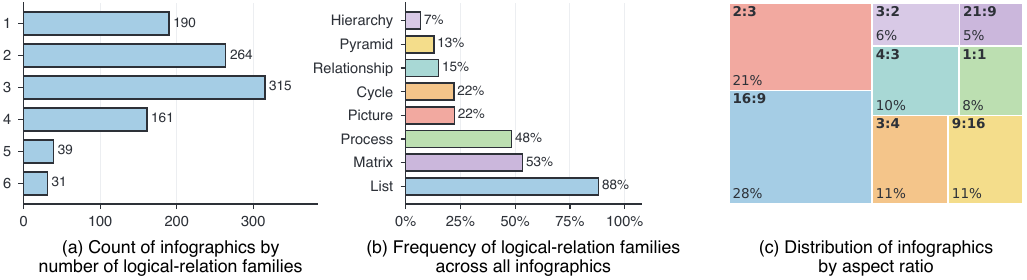}
    \vspace{-1.5em}
    \caption{Distributional summary of \ours. (a) Distinct logical-relation families per infographic; $19\%$ contain one, $81\%$ contain two to six. (b) Fraction of infographics containing each family; an infographic can contain multiple families. (c) Source aspect-ratio distribution over eight ratios.}
    \vspace{-1.5em}
    \label{fig:dataset_stats}
\end{figure*}

\begin{itemize}[itemsep=2pt, parsep=0pt, topsep=0pt, partopsep=0pt, leftmargin=*]
    \item \textit{Stage 1: Topic sourcing.} We define $31$ topic categories where infographics are commonly produced (e.g., companies and athletes), and prompt Claude-Opus-4.7, Gemini-3.1-Pro, and GPT-5.4 to iteratively enumerate $200$ concrete topics per category, yielding $6{,}200$ candidate topics.
    
    \item \textit{Stage 2: Visual reference collection.} For each candidate topic, we generate a reference image using Nanobanana-Pro and GPT-Image-1.5. These references are used only as color palette, not as content to reproduce. Expert annotators then select $1{,}000$ references with aesthetically coherent and topic-faithful content.
    
    \item \textit{Stage 3: Logical template construction.} We build a library of $79$ templates instantiating the eight logical relation families. Each template has two implementations: an HTML version exposing nodes, edges, and containers as addressable DOM elements, and a PowerPoint version using named vector shapes and connectors.
    
    \item \textit{Stage 4: Infographic composition.} Each infographic is composed by transferring the color palette of a reference image onto logical-relation templates. The HTML track is model-assisted: an annotator selects several templates and assembles them together. Claude-Opus-4.7 then performs two passes conditioned on the reference image and selected templates: a styling pass that adapts the CSS to match the reference's color palette, and a content pass that fills textual placeholders with topic-relevant text. The annotator then post-edits the result. In contrast, the PowerPoint track is authored fully manually by annotators, who select and arrange templates and adjust the color palette to match the reference image.
    
    \item \textit{Stage 5: Quality control.} Every infographic is reviewed carefully by a second annotator against three criteria: (i)~structural integrity, no overflow or broken connectors, and complete logical-relation instances; (ii)~stylistic faithfulness, palette matching the reference; and (iii)~content completeness, all textual fields and labels are topic-relevant and role-compatible.
\end{itemize}

\paragraph{Diversity and Coverage.}
Fig.~\ref{fig:dataset_stats} shows that \ours spans the diversity of 
real-world infographics along three axes: \textit{structural 
complexity}, \textit{family coverage}, and \textit{canvas shape}.

\begin{itemize}[itemsep=2pt, parsep=0pt, topsep=0pt, partopsep=0pt, leftmargin=*]
    \item \textit{Structural complexity} (Fig.~\ref{fig:dataset_stats}a). Each infographic contains four blocks on average, drawn from one or more logical-relation families. Only $19\%$ realize a single family; the remaining $81\%$ compose two or more on the canvas, with three families the modal case and instances reaching as many as six. The benchmark therefore exercises both within-family and across-family reflow.

    \item \textit{Family coverage} (Fig.~\ref{fig:dataset_stats}b). All eight families appear, with a non-uniform distribution: List, Matrix, and Process are more frequent, while Hierarchy appears less often. These family annotations are non-mutually exclusive: for example, the $88\%$ reported for List means that $88\%$ of infographics contain a List structure, not that $88\%$ of the dataset belongs exclusively to a List class. List-based organization is a fundamental structure that commonly co-occurs with other relations in infographics. We keep this natural skew rather than enforcing uniform coverage while still including all relation families; per-family results in Appx.~\ref{appx:fine_grained} show that the main model comparison holds across all eight families.
    
    \item \textit{Canvas shape} (Fig.~\ref{fig:dataset_stats}c). Source aspect ratios span landscape ($21$:$9$, $16$:$9$, $4$:$3$, $3$:$2$), portrait ($9$:$16$, $3$:$4$, $2$:$3$), and square ($1$:$1$) formats, allowing evaluation across diverse practical canvas shapes.

\end{itemize}

\begin{table*}[t]
\centering
\setlength{\tabcolsep}{5pt}
\renewcommand{\arraystretch}{0.85}
\resizebox{\linewidth}{!}{%
\begin{tabular}{l l l r c c l}
\toprule
\textbf{Benchmark} & \textbf{Task} & \textbf{Domain} & \textbf{\#Samples} & \makecell{\textbf{Logical}\\\textbf{Relation}} & \makecell{\textbf{Reflow-}\\\textbf{Aware}} & \textbf{Evaluation Metric} \\
\midrule
\multicolumn{7}{l}{\textit{Image Editing}} \\
\midrule
ImgEdit~\cite{ye2025imgedit}      & Image Editing & Natural      & $811$ & \xmark & \xmark & MLLM-judge \\
WiseEdit~\cite{pan2025wiseedit}   & Image Editing & Natural      & $1{,}220$ & \xmark & \xmark & MLLM-judge \\
GEditBench-v2~\cite{jiang2026geditbench}   & Image Editing & Natural  & $1{,}200$ & \xmark & \xmark & Pairwise MLLM \\
\midrule
\multicolumn{7}{l}{\textit{Infographics Generation}} \\
\midrule
BizGen~\cite{peng2025bizgen}        & Text-to-Image  & Infographic  & $1{,}000$ & \xmark & --     & MLLM + OCR \\
BizGenEval~\cite{li2026bizgeneval}  & Text-to-Image  & Infographic   & $400$ & \xmark & --     & MLLM-judge \\
IGenBench~\cite{tang2026igenbench}  & Text-to-Image  & Infographic  & $600$ & \xmark & --     & MLLM-judge \\
\midrule
\rowcolor{gray!15}
\ours (Ours) & Image Editing & Infographic & $4{,}000$ & \cmark & \cmark & MLLM-judge \\
\bottomrule
\end{tabular}%
}
\caption{Comparison of \ours with related benchmarks. ``--'' indicates not applicable to text-to-image generation. The $4{,}000$ samples come from $1{,}000$ infographics with $4$ edit tasks each.}
\label{tab:benchmark_comparison}
\vspace{-1.5em}
\end{table*}

\subsection{Editing Tasks}
\label{sec:editing_tasks}
As illustrated in Fig.~\ref{fig:framework} (middle), \ours defines four editing tasks, ordered by the scope of structural perturbation they demand, from text-level changes to canvas-level restructurings:

\begin{itemize}[itemsep=2pt, parsep=0pt, topsep=0pt, partopsep=0pt, leftmargin=*]
    \item \textsc{Expand-Text} lengthens the text of one target element, forcing the host to grow while its neighbors absorb the displacement through reflow.
    \item \textsc{Insert-Element} adds one or more new elements into an existing logical-relation block (e.g., a step in a Process), so the host structure must reflow, such as re-routing connectors, and redistributing space among existing siblings.
    \item \textsc{Swap-Block} exchanges the canvas positions of two logical-relation blocks as whole units, requiring the rest of the canvas to reflow by resizing and re-aligning around the new arrangement.
    \item \textsc{Reshape-Canvas} re-renders the infographic at a different aspect ratio, demanding a canvas-level reflow in which every block is re-positioned, re-scaled, and re-aligned to fit the new shape.
\end{itemize}

\paragraph{Instruction Annotation.}
For each of the $1{,}000$ infographics, annotators write one instruction for each editing task, yielding $4{,}000$ (infographic, instruction) pairs. We require every instruction to be feasible and reflow-inducing, as verified by three checks. (i)~\emph{Spatial coupling}: targets are placed in regions where the edit affects nearby nodes, cards, text fields, or connectors, forcing displacement, resizing, re-wrapping, re-alignment, or rerouting. (ii)~\emph{Shape-sensitive anchoring}: targets are drawn from shaped or connected relation instances, such as pyramid tiers or hierarchy branches, to test diverse geometry, family-specific spacing, and connector structure. (iii)~\emph{Coverage}: the four instructions for each infographic cover different regions and different relation families.

\paragraph{Difficulty Levels.}
\ours defines three difficulty levels for tasks with explicit magnitude
axes: \textsc{Expand-Text}, \textsc{Insert-Element}, and
\textsc{Swap-Block}. Easy, medium, and hard correspond to $5$--$10$,
$11$--$20$, and $21$--$30$ appended words for \textsc{Expand-Text}; $1$,
$2$--$3$, and $4$--$5$ inserted elements for \textsc{Insert-Element}; and
$1$, $2$, and $3$ swapped block pairs for \textsc{Swap-Block}.
\textsc{Reshape-Canvas} has no discrete magnitude axis. Full difficulty-level statistics
are provided in Appx.~\ref{appx:difficulty_statistics}.

\subsection{Comparison with Existing Benchmarks}
\label{sec:comparison}

Tab.~\ref{tab:benchmark_comparison} contrasts \ours with related benchmarks.
Existing image-editing benchmarks~\cite{ye2025imgedit, pan2025wiseedit, jiang2026geditbench}
focus on natural images, often assume pixel-level preservation of unmentioned
regions, and do not emphasize logical-relation taxonomies, leaving reflow outside their task design and evaluation. Infographic generation
benchmarks~\cite{peng2025bizgen, li2026bizgeneval, tang2026igenbench}
share our domain, but study text-to-image generation, and do not systematically account for logical-relation diversity. In contrast, \ours first targets infographic editing, organizes samples by logical-relation families, and evaluates reflow-aware edits.

\section{Evaluation Protocol}
\label{sec:eval}

We operationalize the compliance and preservation predicates of Sec.~\ref{sec:task} as two binary aspects: \emph{Edit Compliance} and \emph{Content Preservation}. For each $(X, I, X')$ triple, we prompt an MLLM with the original image, edited image, and instruction, using a task-specific rubric that reflects the edit's reflow contract. The judge returns a JSON response with binary verdicts for the two aspects and an overall verdict equal to their conjunction. The four tasks share this schema but use task-specific criteria; full prompts are provided in Appx.~\ref{appx:prompts}.

\begin{itemize}[itemsep=2pt, parsep=0pt, topsep=0pt, partopsep=0pt, leftmargin=*]
    \item \textit{Edit Compliance (EC).} \textsc{Expand-Text} requires the named target to carry a moderately longer, on-topic body text fully rendered inside the canvas. \textsc{Insert-Element} requires the new element to appear at the location named in the instruction with its content fully rendered. \textsc{Swap-Block} requires the two named blocks to exchange canvas locations as whole units, each carrying its content unchanged. \textsc{Reshape-Canvas} requires the canvas to match the target aspect ratio.
    \item \textit{Content Preservation (CP).} Since \ours instructions are reflow-inducing, CP does not require pixel-level immobility. Instead, it checks whether unedited source content remains intact after necessary reflow, such as repositioning, resizing, re-alignment, and connector rerouting.
    
\end{itemize}
We report \emph{Success Rate (SR)} as the percentage of edits for which both \emph{Edit Compliance} and \emph{Content Preservation} are judged successful.

\begin{table*}[t]
\centering
\small
\setlength{\tabcolsep}{6pt}
\renewcommand{\arraystretch}{1.0}
\begin{tabular}{@{}l rrr rrr rrr rrr r@{}}
\toprule
& \multicolumn{3}{c}{\textbf{\textsc{Expand-Text}}} 
& \multicolumn{3}{c}{\textbf{\textsc{Insert-Element}}} 
& \multicolumn{3}{c}{\textbf{\textsc{Swap-Block}}}
& \multicolumn{3}{c}{\textbf{\textsc{Reshape-Canvas}}}
& \multirow{2}{*}{\textbf{Avg.}} \\
\cmidrule(lr){2-4} \cmidrule(lr){5-7} \cmidrule(lr){8-10} \cmidrule(lr){11-13}
\textbf{Model} & \multicolumn{1}{c}{EC} & \multicolumn{1}{c}{CP} & \multicolumn{1}{c}{SR} & \multicolumn{1}{c}{EC} & \multicolumn{1}{c}{CP} & \multicolumn{1}{c}{SR} & \multicolumn{1}{c}{EC} & \multicolumn{1}{c}{CP} & \multicolumn{1}{c}{SR} & \multicolumn{1}{c}{EC} & \multicolumn{1}{c}{CP} & \multicolumn{1}{c}{SR} & \\
\midrule
\multicolumn{14}{c}{\textit{Proprietary}} \\
\midrule
GPT-Image-2        & 99.0 & 86.7 & 85.9 & 85.4 & 80.7 & 72.0 & 29.0 & 79.2 & 25.5 & 96.3 & 65.1 & 64.4 & 62.0 \\
Nanobanana-Pro     & 87.1 & 64.0 & 57.8 & 58.8 & 57.8 & 40.2 & 36.5 & 79.8 & 32.0 & 87.2 & 36.9 & 36.4 & 41.6 \\
Nanobanana-2       & 85.1 & 58.6 & 52.6 & 65.3 & 49.4 & 39.0 & 30.0 & 72.1 & 24.0 & 92.0 & 36.4 & 36.3 & 38.0 \\
Nanobanana         & 23.2 & 23.8 &  9.2 &  3.1 & 14.6 &  0.4 &  1.6 & 41.8 &  0.9 & 25.4 &  8.4 &  3.5 &  3.5 \\
Seedream-5.0-Lite  & 22.1 & 28.0 &  9.1 & 12.3 & 11.0 &  1.3 &  9.3 & 37.0 &  5.2 & 64.1 &  4.6 &  4.3 &  5.0 \\
\midrule
\multicolumn{14}{c}{\textit{Open-Weight}} \\
\midrule
HunyuanImage-3.0 & 21.3 & 34.8 & 12.6 &  2.5 &  4.3 &  0.0 &  0.4 & 16.4 &  0.0 &  1.4 &  5.2 &  0.0 &  3.2 \\
Qwen-Image-Edit      &  9.2 &  3.3 &  1.5 &  1.2 &  2.7 &  0.0 &  0.0 &  6.3 &  0.0 & 16.5 &  0.0 &  0.0 &  0.4 \\
FLUX.2-klein-base-9B      &  6.1 &  7.2 &  0.0 &  0.0 & 21.2 &  0.0 &  0.0 & 77.8 &  0.0 & 20.3 &  0.0 &  0.0 &  0.0 \\
\bottomrule
\end{tabular}
\caption{Editing performance of eight image-editing models on \ours across four editing tasks. EC = edit compliance, CP = content preservation, SR = success rate, the conjunction of both. All values are percentages. Avg.\ is the mean SR across tasks.}
\label{tab:main_results}
\vspace{-1.5em}
\end{table*}
\section{Experiments}
\label{sec:experiments}

\subsection{Setup}
We benchmark eight  frontier image-editing models spanning proprietary and open-source families. The proprietary models include GPT-Image-2~\cite{openai2026gptimage2}, Nanobanana-Pro~\cite{google2025gemini3image}, Nanobanana-2~\cite{google2026nanobanana2}, Nanobanana~\cite{google2025nanobanana}, and Seedream-5.0-Lite~\cite{seedream2026seedream5lite}. The open-source models include Qwen-Image-Edit~\cite{wu2025qwenimage}, HunyuanImage-3.0~\cite{cao2025hunyuanimage}, and FLUX.2-klein-base-9B~\cite{bfl2026flux2klein}. All models are queried with our $4{,}000$ (infographic, instruction) pairs under each model's recommended inference settings, and every triple is judged by Gemini-3.1-Pro~\cite{google2026gemini31pro}, validated against expert human annotators in Sec.~\ref{sec:human_agreement}.

\subsection{Main Results}
\label{sec:main_results}

\paragraph{A clear capability gap separates the frontier proprietary models from the rest.}~As shown in Tab.~\ref{tab:main_results}, GPT-Image-2 leads at $62.0\%$ average SR, followed by Nanobanana-Pro and Nanobanana-2 at $41.6\%$ and $38.0\%$. The remaining five models, including Nanobanana, Seedream-5.0-Lite, and all open-weight models, fall at or below $7\%$. The gap reflects more than a gradual quality difference: many models fail to produce usable infographics at all, underscoring the need for \ours.

\paragraph{Failure modes are task-specific.}
The EC/CP decomposition shows that each edit task stresses a different part of reflow. On \textsc{Expand-Text}, the top three proprietary models largely execute the target edit, with EC between $85.1\%$ and $99.0\%$, but SR still trails EC because CP remains lower. \textsc{Reshape-Canvas} exposes a stronger preservation bottleneck: the same models achieve high EC ($87.2$--$96.3\%$), yet CP drops to $36.4$--$65.1\%$, indicating that canvas-level reflow often damages source content or visual structure. \textsc{Insert-Element} is difficult on both axes, yielding $72.0\%$ SR for GPT-Image-2 but only around $39$--$40\%$ for the two Nanobanana variants. \textsc{Swap-Block} is bottlenecked by edit compliance: no model exceeds $36.5\%$ EC or $32.0\%$ SR, while CP can remain comparatively high, showing that compliance and preservation decouple on block-level swaps.

\paragraph{Open-weight models lag behind by a wide margin.}
HunyuanImage-3.0, Qwen-Image-Edit, and FLUX.2-klein-base-9B reach $3.2\%$, $0.4\%$, and $0.0\%$ average SR, all far below the strongest proprietary model GPT-Image-2 ($62.0\%$). The low averages are not driven by a single failure task: all three open-weight models score $0\%$ SR on \textsc{Insert-Element}, \textsc{Swap-Block}, and \textsc{Reshape-Canvas}. High CP alone does not imply success: FLUX.2-klein-base-9B reaches $77.8\%$ CP on \textsc{Swap-Block} but $0.0\%$ EC, so the requested structural edit is not completed. These results reveal a substantial gap between open-weight and proprietary editors, and \ours identifies reflow as a concrete target for improving open-weight models.
\begin{table*}[!t]
\centering
\small
\setlength{\tabcolsep}{6pt}
\renewcommand{\arraystretch}{0.85}
\resizebox{\textwidth}{!}{%
\begin{tabular}{@{}l rrr rrr rrr rrr r@{}}
\toprule
& \multicolumn{3}{c}{\textbf{\textsc{Expand-Text}}}
& \multicolumn{3}{c}{\textbf{\textsc{Insert-Element}}}
& \multicolumn{3}{c}{\textbf{\textsc{Swap-Block}}}
& \multicolumn{3}{c}{\textbf{\textsc{Reshape-Canvas}}} 
& \multicolumn{1}{c}{\multirow{2}{*}{\textbf{Avg.}}} \\
\cmidrule(lr){2-4} \cmidrule(lr){5-7} \cmidrule(lr){8-10} \cmidrule(lr){11-13}
\textbf{Model} & \multicolumn{1}{c}{EC} & \multicolumn{1}{c}{CP} & \multicolumn{1}{c}{SR} 
& \multicolumn{1}{c}{EC} & \multicolumn{1}{c}{CP} & \multicolumn{1}{c}{SR} 
& \multicolumn{1}{c}{EC} & \multicolumn{1}{c}{CP} & \multicolumn{1}{c}{SR} 
& \multicolumn{1}{c}{EC} & \multicolumn{1}{c}{CP} & \multicolumn{1}{c}{SR} & \\
\midrule
\multicolumn{14}{c}{\textit{Code}} \\
\midrule
Gemini-3.1-Pro    & 85.7 & 72.3 & 67.4 & 85.6 & 70.7 & 68.0 & 68.8 & 77.6 & 59.6 & 69.4 & 45.2 & 45.1 & 60.0 \\
Gemini-3.5-Flash  & 83.3 & 68.1 & 64.1 & 84.1 & 64.5 & 62.0 & 60.1 & 79.2 & 54.8 & 70.4 & 42.4 & 42.2 & 55.8 \\
Gemini-3.1-Flash  & 79.6 & 65.5 & 60.6 & 63.4 & 49.1 & 40.2 & 56.1 & 72.7 & 47.1 & 38.0 &  5.3 &  5.2 & 38.3 \\
\midrule
\multicolumn{14}{c}{\textit{Code\,+\,Image}} \\
\midrule
Gemini-3.1-Pro    & 88.2 & 72.0 & 68.5 & 85.3 & 72.2 & 71.0 & 69.0 & 78.4 & 60.2 & 71.8 & 47.2 & 46.8 & 61.6 \\
Gemini-3.5-Flash  & 85.1 & 69.5 & 65.8 & 85.4 & 67.2 & 63.2 & 60.9 & 79.8 & 56.8 & 71.2 & 46.6 & 43.1 & 57.2 \\
Gemini-3.1-Flash  & 80.2 & 64.5 & 59.6 & 64.3 & 51.4 & 43.0 & 56.4 & 74.0 & 50.2 & 38.1 &  8.7 &  8.6 & 40.4 \\
\bottomrule
\end{tabular}%
}
\caption{Code-level editing on \ours. 
\emph{Code} feeds the source code alone; \emph{code\,+\,image} 
adds the rendered image. EC = edit compliance, CP = content 
preservation, SR = the conjunction of both. All values are 
percentages. Avg.\ is the mean SR across tasks.}
\label{tab:code_vs_image}
\end{table*}
\section{Discussion}
\label{sec:discussion}
\subsection{Pixel-Level versus Code-Level Editing}
\label{sec:code_generation}

Structured visual content generation and editing remain open challenges for multimodal systems, and can be approached through two pathways~(detailed in Appx.~\ref{appx:related_work}): pixel-level pathway with image generation models, or code-level pathway over textual source. Since every infographic in \ours is implemented with HTML or PowerPoint, we can directly compare code-level editing with pixel-level image editing under the same evaluation protocol. We test two input modalities: 
\textit{Code} feeds the model the source code alone, while 
\textit{Code + Image} additionally provides the rendered original 
as visual grounding. Gemini-3.1-Pro, Gemini-3.5-Flash and Gemini-3.1-Flash serve as 
code generators, and we render their edited outputs into images.

As shown in Tab.~\ref{tab:code_vs_image}, Gemini-3.1-Pro with \textit{Code + Image} reaches $61.6\%$ average SR, just matching GPT-Image-2's $62.0\%$ average SR. However, the task-level pattern is not uniform. GPT-Image-2 remains stronger on \textsc{Expand-Text} and \textsc{Reshape-Canvas}, while Gemini-3.1-Pro is comparable on \textsc{Insert-Element} and much stronger on \textsc{Swap-Block}, improving EC from $29.0\%$ to $69.0\%$ and SR from $25.5\%$ to $60.2\%$. Source code makes named blocks directly addressable, turning block swapping into an explicit structural operation. These findings reveal complementary strengths between the two pathways and point toward future improvements, or even hybrid systems, that combine their respective advantages.

\subsection{Interactive Selection Hint}
\label{sec:interactive_selection}

\begin{table}[t]
\centering
\small
\setlength{\tabcolsep}{4pt}
\renewcommand{\arraystretch}{0.85}
\begin{tabular}{@{}llcccc@{}}
\toprule
\textbf{Model} & \textbf{Hint} & EC & CP & SR & $\Delta$SR \\
\midrule
\multirow{2}{*}{GPT-Image-2}
  & w/o & 29.0 & 79.2 & 25.5 & -- \\
  & w/  & 34.2 & 92.7 & 32.9 & +7.4 \\
\midrule
\multirow{2}{*}{Nanobanana-Pro}
  & w/o & 36.5 & 79.8 & 32.0 & -- \\
  & w/  & 39.4 & 83.6 & 35.2 & +3.2 \\
\midrule
\multirow{2}{*}{Nanobanana-2}
  & w/o & 30.0 & 72.1 & 24.0 & -- \\
  & w/  & 34.2 & 79.1 & 27.5 & +3.5 \\
\bottomrule
\end{tabular}
\caption{Effect of interactive selection hints on \textsc{Swap-Block} performance. 
}
\label{tab:interactive_selection}
\end{table}

\textsc{Swap-Block} failures could stem from two distinct causes: 
the model fails to execute the structural verb (\emph{swap}), or 
it may just fail to resolve which two blocks the instruction names.
To separate these, we annotate the input 
image with interactive selection hints: each swap pair is marked 
with bounding boxes and pair indices~(detailed in Appx.~\ref{app:interactive-selection-hints}), and the model is 
instructed to treat the marks as localization guidance. If the failure were primarily a 
target-localization problem, hints should produce a large gain; 
if the failure is genuinely about reflow, the gain should be 
modest.

Tab.~\ref{tab:interactive_selection} reports the result. The gain 
is consistently modest: $+3.2$ to $+7.4$ SR across the three 
strongest editors, with GPT-Image-2 benefiting most ($25.5 
\rightarrow 32.9$) and the two NanoBanana variants gaining only 
$3$--$3.5$ points. Even with perfect localization, no model 
exceeds $36\%$ SR on \textsc{Swap-Block}. The bottleneck \ours surfaces is therefore not simply the inability to localize the target blocks; it is the reflow operation itself, which current editors cannot reliably perform even when the targets are unambiguously marked.

\subsection{Different Difficulty Levels}

\begin{figure}[t]
    \centering
    \includegraphics[width=\linewidth]{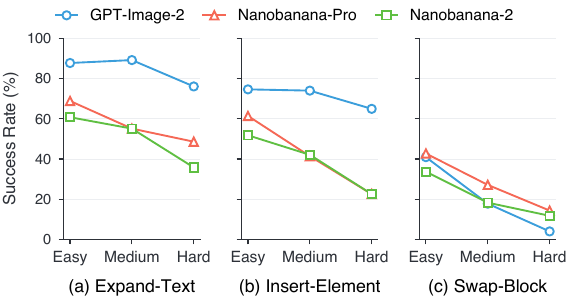}
    \caption{Success rates across difficulty levels for edit tasks: \textsc{Expand-Text}, \textsc{Insert-Element}, and \textsc{Swap-Block}.}
    \label{fig:difficulty}
\end{figure}

Fig.~\ref{fig:difficulty} breaks down performance by difficulty level for
the three tasks with explicit magnitude axes: \textsc{Expand-Text},
\textsc{Insert-Element}, and \textsc{Swap-Block}. Across all three tasks,
success rates generally decrease as requested edits become harder,
confirming that \ours captures graded difficulty. The degradation is
mildest for \textsc{Expand-Text}, where top models can often absorb longer
text by shrinking fonts or locally re-wrapping neighboring elements. In
contrast, \textsc{Insert-Element} shows a sharper drop, especially for
weaker models, because adding multiple elements requires redistributing
existing space and maintaining logical-relation consistency.
\textsc{Swap-Block} is the most difficult across all settings: even at the
easy level, models achieve substantially lower success rates than on the
other two tasks, and performance falls quickly as more block pairs are
exchanged.

\begin{figure*}[!t]
    \centering
    \includegraphics[width=\textwidth]{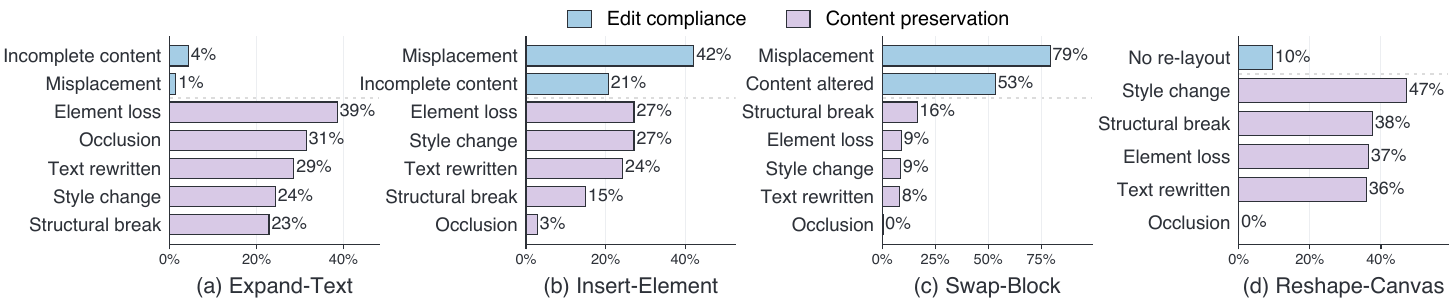} 
    \caption{Distribution of failure types per task. We categorize unsuccessful cases by the diagnostic sub-field they violate, separated into Edit Compliance (blue) and Content Preservation (purple). Detailed error cases in Appx.~\ref{appx:case_study}.}
    \label{fig:error_analysis}
\end{figure*}

\begingroup
\subsection{Human Agreement and Judge Robustness}
\label{sec:human_agreement}
To validate our MLLM-as-a-judge protocol, three expert annotators independently evaluate $800$ outputs from each of GPT-Image-2, Nanobanana-2, and HunyuanImage-3.0, balanced across the four editing tasks. 
Their majority verdict serves as the human reference. Gemini-3.1-Pro achieves agreement of $0.920$ to $0.981$ across the three models.
\looseness=-1
Agreement also remains high when stratified by task, difficulty, and predicted outcome, and is stable under a cross-vendor judge and repeated evaluation. Full results are reported in Appx.~\ref{appx:human_evaluation}.

\endgroup

\subsection{Error Analysis}
We decompose each failure case from GPT-Image-2, the strongest editor we evaluate, into the diagnostic sub-fields of Sec.~\ref{sec:eval} and report the frequency of each failure type in Fig.~\ref{fig:error_analysis}. Three patterns emerge.

First, \textsc{Misplacement} dominates the structurally demanding tasks, accounting for $79\%$ of \textsc{Swap-Block} failures and $42\%$ of \textsc{Insert-Element} failures. The model can identify 
that a swap or insertion is required, but cannot place the named block or element correctly through reflow; this matches the finding in Sec.~\ref{sec:interactive_selection} that interactive selection hints only partially relieve. 

Second, \textsc{Expand-Text} exhibits a different pattern: edit compliance rarely fails ($\leq 4\%$), while most errors arise from content preservation failures. The target text is often expanded correctly, but nearby nodes fail to reflow, causing element loss, occlusion or structural break. This silent-damage pattern highlights the need to evaluate content preservation separately from edit compliance.

Third, \textsc{Reshape-Canvas} concentrates its failures on global stylistic and structural breakdown ($47\%$ style change, $38\%$ structural break), indicating that re-rendering at a new aspect ratio rarely preserves the original visual organization. Even when the canvas reaches the target ratio, the layout inside is re-painted, losing the source palette and structural relation. Occlusion is the most common preservation failure on \textsc{Expand-Text} at $31\%$, but drops to $0\%$ on 
\textsc{Swap-Block} and \textsc{Reshape-Canvas}, where elements redistribute globally rather than overwhelm a neighbor.

Overall, the failures show that reflow is multi-faceted rather than a single operation: \textsc{Insert-Element} and \textsc{Swap-Block} stress correct placement, \textsc{Expand-Text} stresses neighbor-aware adjustment, and \textsc{Reshape-Canvas} stresses global style and structure preservation.

\section{Related Work}
\paragraph{Image Editing Benchmarks.}
Existing image-editing benchmarks largely follow a pixel-identity
assumption. ImgEdit-Bench~\cite{ye2025imgedit} evaluates natural-image
edits on instruction adherence, editing quality, and detail preservation,
while WiseEdit~\cite{pan2025wiseedit} introduces more cognition- and
creativity-informed edits. Other benchmarks and editing
datasets~\cite{zhang2023magicbrush, brooks2023instructpix2pix,hui2024hqedit, ge2024seeddataedit,
zhao2024ultraedit, ma2024i2ebench,yu2025anyedit,jia2025compbench,gao2026text,zhao2026envisioning} differ in scale, operation taxonomy,
and annotation cost, but similarly treat unmentioned regions as content to
preserve unchanged. 
This pixel-identity assumption is reasonable for natural photographs, but breaks down on structured content, where edits often require layout changes.
These benchmarks, therefore, do not target structured visual content editing or evaluate the required reflow capability.

\paragraph{Structured Visual Content Generation.} 
Current multimodal foundation models have demonstrated strong capabilities in generating structured visual content such as infographics~\cite{google2025stitch,yang2025chartmimic,NotebookLM,zhu2026paperbanana,zhu2026autofigure}, and recent benchmarks evaluate this regime: BizGen~\cite{peng2025bizgen} on visual text rendering for infographics and slides, IGenBench~\cite{tang2026igenbench} on text-to-infographic reliability through atomic yes/no questions, and BizGenEval~\cite{li2026bizgeneval} on diverse commercial document generation.  These works focus on creating structured visuals from text, whereas \ours studies editing existing infographics with diverse logical relations while preserving their structure through reflow.

\section{Conclusion}
\label{sec:conclusion}

We present \ours, a novel benchmark for infographic editing that probes
\emph{reflow}, a capability for which current image-editing models are
not systematically evaluated.
\ours covers $1{,}000$ infographics across eight 
logical-relation families and $4{,}000$ instruction-paired editing 
tasks, together with a reflow-aware evaluation protocol that 
decomposes each verdict into edit compliance and content 
preservation. Across eight frontier editors, only GPT-Image-2 
clears $60\%$ average success rate, and no model exceeds $36\%$ 
on \textsc{Swap-Block} even when given perfect target localization, 
identifying reflow as the dominant bottleneck.
We further examine code-level editing and conduct error analyses to
provide additional diagnostic insights.
We anticipate that \ours will motivate training and evaluation
frameworks that treat reflow as a central capability of structured visual
content editing.

\section*{Limitations}
\ours evaluates single-turn editing instructions to cleanly attribute each outcome to a specific editing task; real-world design workflows may involve multi-turn refinements, where users iteratively adjust content, layout, and style. Extending reflow evaluation to such interactive settings is another promising direction.

\section*{Acknowledgements}
This research was supported in part by the NSF under grants 2146151 and 2422214.

\bibliography{references}

\etocdepthtag.toc{appendixtag}

\clearpage

\appendix

\etocsettagdepth{mainmatter}{none}
\etocsettagdepth{appendixtag}{subsection}
\etocsettocstyle{\section*{Appendix}}{}
\tableofcontents

\clearpage

\section{Details on Benchmark Construction}
\label{appx:annotation_pipeline}

\subsection{Infographics Construction}
\label{appx:infographics_construction}

Existing infographic resources~\cite{mathew2022infographicvqa,li2025chartgalaxy,zhu2025infodet,xie2026infochartqa,tang2026igenbench,zhang2026pixelcraft} do not satisfy the requirements of infographic editing evaluation.
First, many datasets provide only rendered images, sometimes with limited visual quality, and lack editable source files.
This prevents code-level editing and limits diagnostic analysis, since target elements and surrounding structures are not explicitly addressable.
Consequently, addressable tasks (detailed in Appx.~\ref{appx:addressable_tasks}) would have to be defined by visual intuition over pixels rather than by concrete operations on editable HTML or PowerPoint objects.
Second, many existing infographic datasets are designed for VQA, OCR, or chart-centric understanding and generation, where the main challenge is reading text, extracting values, or interpreting charts.
They therefore offer limited structural diversity and do not systematically cover logical relations.
We thus build \ours{} from logical-relation templates and editable sources, so that each sample supports feasible, non-trivial, and reflow-inducing edits.

\subsubsection{Stage 1: Topic Sourcing} 
    We define $31$ topic categories spanning domains in which infographics are 
    commonly produced (e.g., companies, products, athletes). To reduce the 
    topical and stylistic bias that may arise from relying on a single language 
    model, we use a multi-model topic sourcing procedure. For each category, 
    Claude-Opus-4.7 first enumerates $70$ concrete topics. We then provide the 
    existing list to Gemini-3.1-Pro and ask it to supplement the category with 
    additional non-duplicate topics. Finally, we provide the accumulated list to 
    GPT-5.4 and ask it to further complete the category until the list reaches 
    $200$ topics. The resulting topics are deduplicated within each category and 
    shuffled before entering the next stage. Across $31$ categories, this process 
    yields $6{,}200$ candidate topics.

\subsubsection{Stage 2: Visual Reference Collection}
    For each candidate topic, we generate a richly designed reference infographic 
    that serves as a reference for color palette. To 
    avoid tying the color palette of the benchmark to a single image generation 
    model, we split the $6{,}200$ candidate topics evenly across two generators, 
    NanoBanana-Pro and GPT-Image-1.5, with each model generating 
    approximately half of the reference images. Five expert annotators 
    independently filter these candidates down to $1{,}000$ references, keeping 
    only those with coherent palettes, and topic-faithful 
    content. They serve as reference that 
    supply concrete guidance on color palette for the rest 
    of the pipeline, sparing annotators from designing color from scratch.
    
\subsubsection{Stage 3: Logical Template Construction}
\begin{table}[t]
\centering
\small
\begin{tabular}{lr@{\hskip 1.5em}lr}
\toprule
Family & \#Templates & Family & \#Templates \\
\midrule
List         & 16 & Matrix       & 6 \\
Process      & 16 & Pyramid      & 6 \\
Relationship & 15 & Hierarchy    & 4 \\
Cycle        & 7  & Picture      & 9 \\
\midrule
\multicolumn{3}{r}{\textbf{Total}} & \textbf{79} \\
\bottomrule
\end{tabular}
\caption{Templates per logical-relation family in the logical template library. Each template is implemented in both 
HTML and PowerPoint.}
\label{tab:diagram_library}
\end{table}

We build a logical template library that instantiates the 
$8$ logical-relation families, comprising 
$79$ templates in total, distributed across families as in 
Table~\ref{tab:diagram_library}. These families abstract common organization
patterns in structured visual content.

\begin{itemize}[itemsep=2pt, parsep=0pt, topsep=0pt, partopsep=0pt, leftmargin=*]
    \item \textsc{List} templates present grouped
or non-sequential items with comparable importance; 
    
    \item \textsc{Process} templates
encode ordered steps, timelines, or directional workflows; 

    \item \textsc{Cycle}
templates represent recurring stages or closed-loop dependencies;

    \item \textsc{Hierarchy} templates express parent--child, reporting, or multi-level
tree relations; 

    \item  \textsc{Relationship} templates capture comparisons,
overlaps, containment, convergence, opposition, or central--peripheral
associations;

    \item \textsc{Matrix} templates organize concepts by quadrants or
two-dimensional axes;

    \item  \textsc{Pyramid} templates convey layered, proportional,
or bottom-up/top-down structures;

    \item \textsc{Picture} templates use image
slots as first-class carriers of information, often paired with captions or
short textual annotations.

\end{itemize}

\begin{figure*}[!t]
    \centering
    \includegraphics[width=\textwidth]{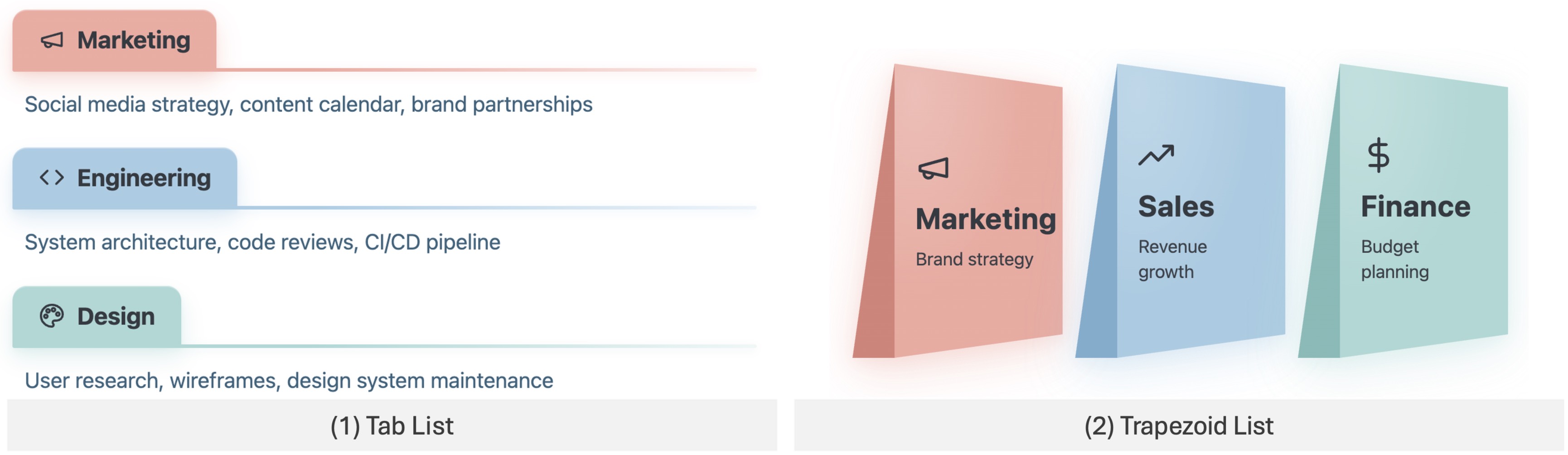} 
    \caption{\textsc{List}: Tab List and Trapezoid List.}
    \label{fig:list}
\end{figure*}

\begin{figure*}[!t]
    \centering
    \includegraphics[width=\textwidth]{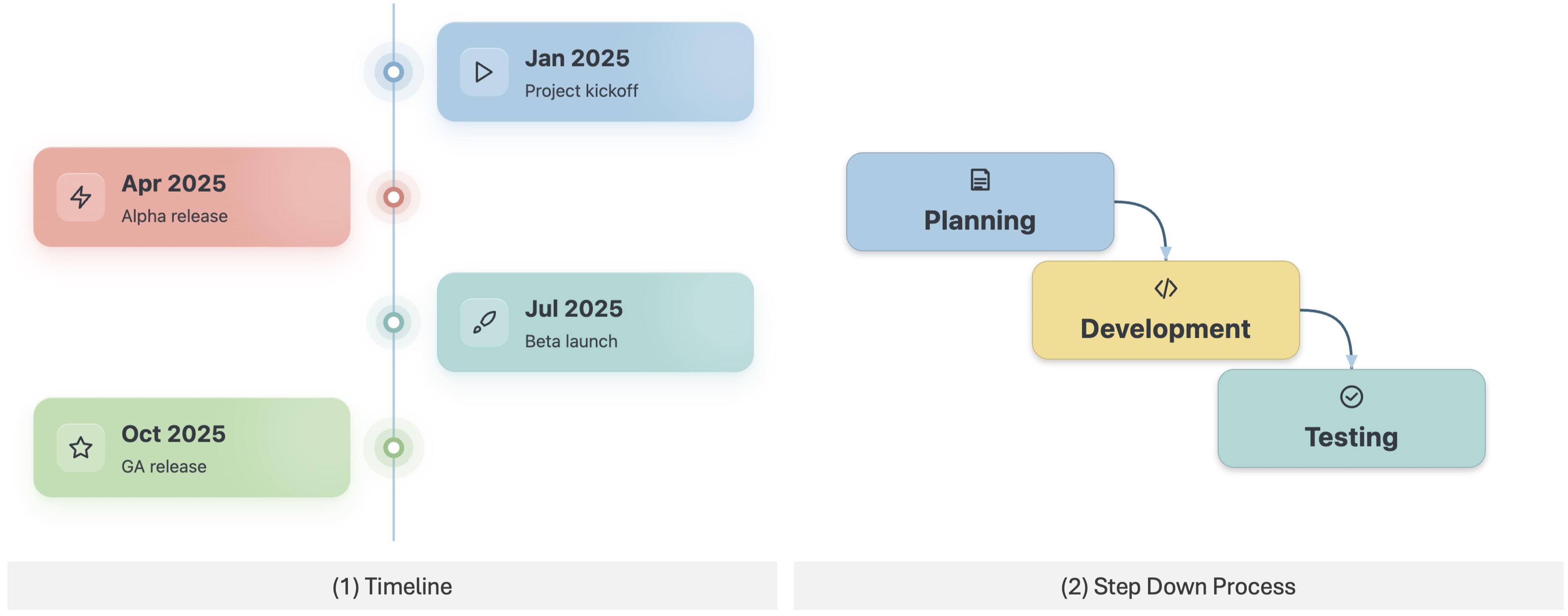} 
    \caption{\textsc{Process}: Timeline and Step Down Process.}
    \label{fig:process}
\end{figure*}

\begin{figure*}[!t]
    \centering
    \includegraphics[width=\textwidth]{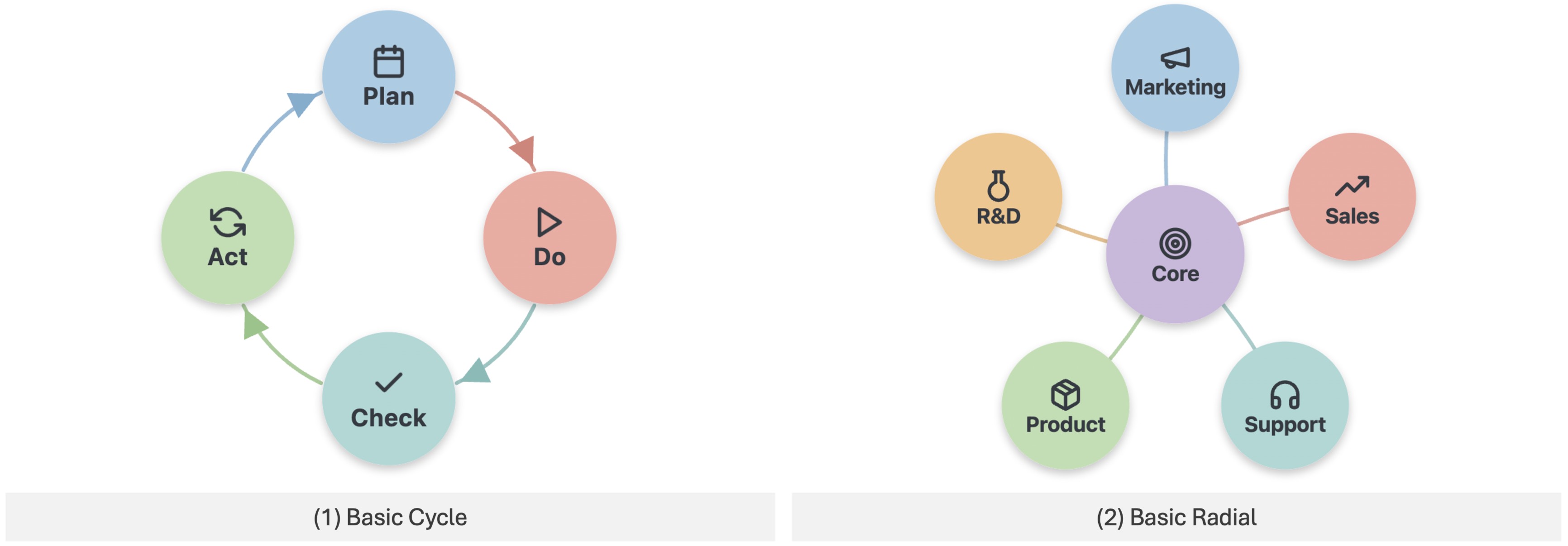} 
    \caption{\textsc{Cycle}: Basic Cycle and Basic Radial.}
    \label{fig:cycle}
\end{figure*}

\begin{figure*}[!t]
    \centering
    \includegraphics[width=\textwidth]{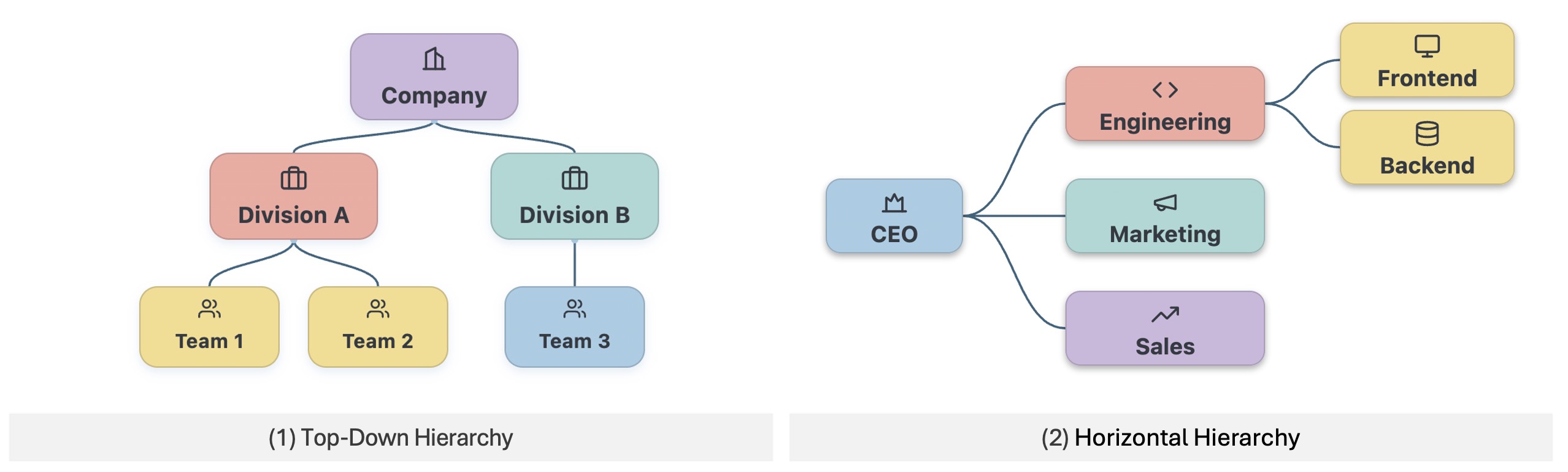} 
    \caption{\textsc{Hierarchy}: Top-Down Hierarchy and Horizontal Hierarchy.}
    \label{fig:hierarchy}
\end{figure*}

\begin{figure*}[!t]
    \centering
    \includegraphics[width=\textwidth]{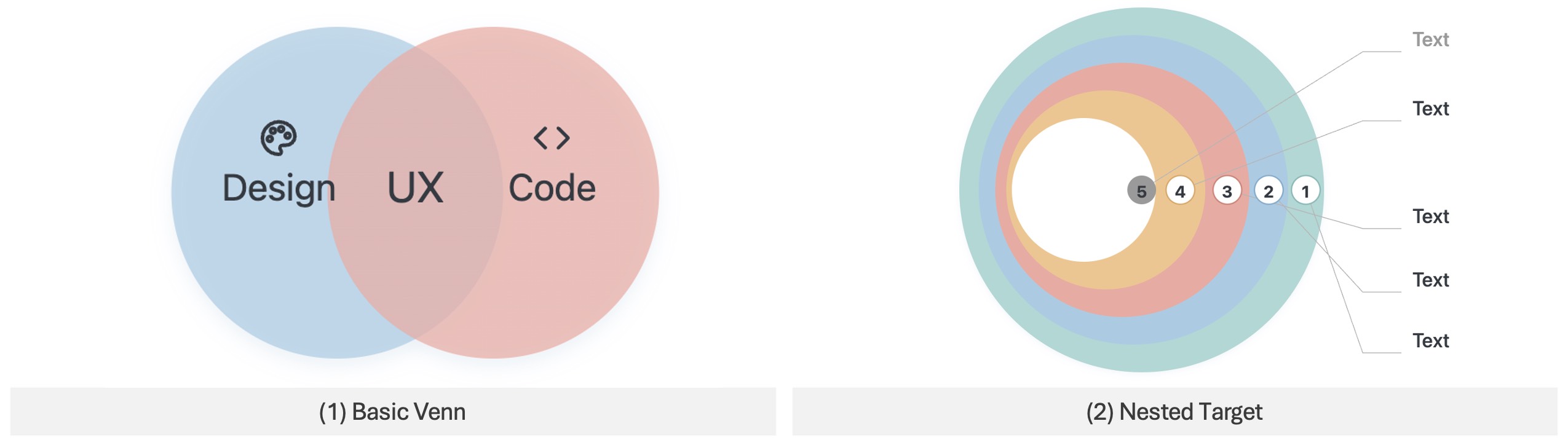} 
    \caption{\textsc{Relationship}: Basic Venn and Nested Target.}
    \label{fig:relationship}
\end{figure*}

\begin{figure*}[!t]
    \centering
    \includegraphics[width=\textwidth]{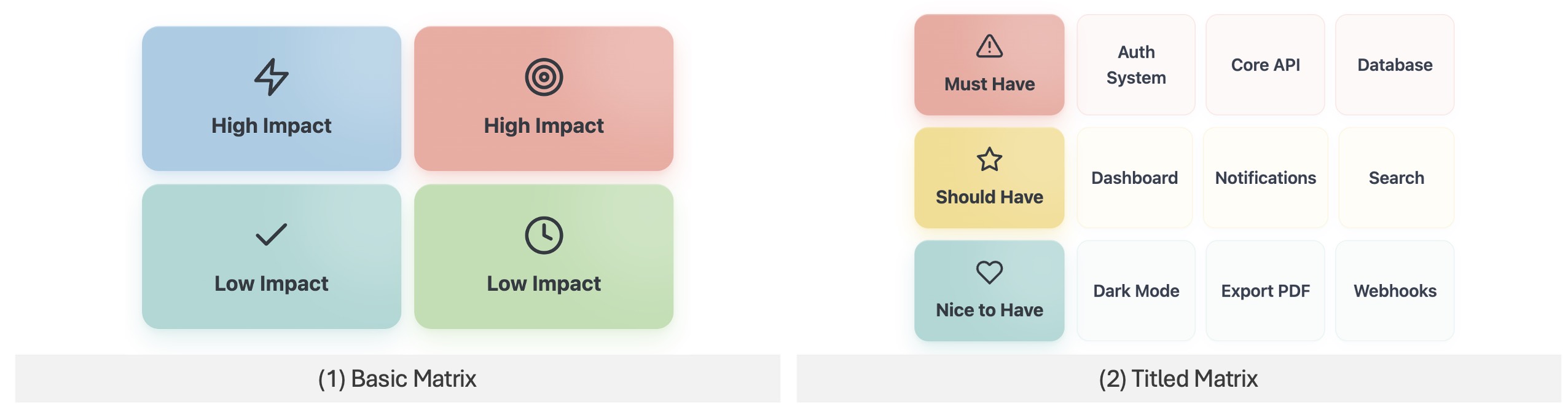} 
    \caption{\textsc{Matrix}: Basic Matrix and Titled Matrix.}
    \label{fig:matrix}
\end{figure*}

\begin{figure*}[!t]
    \centering
    \includegraphics[width=\textwidth]{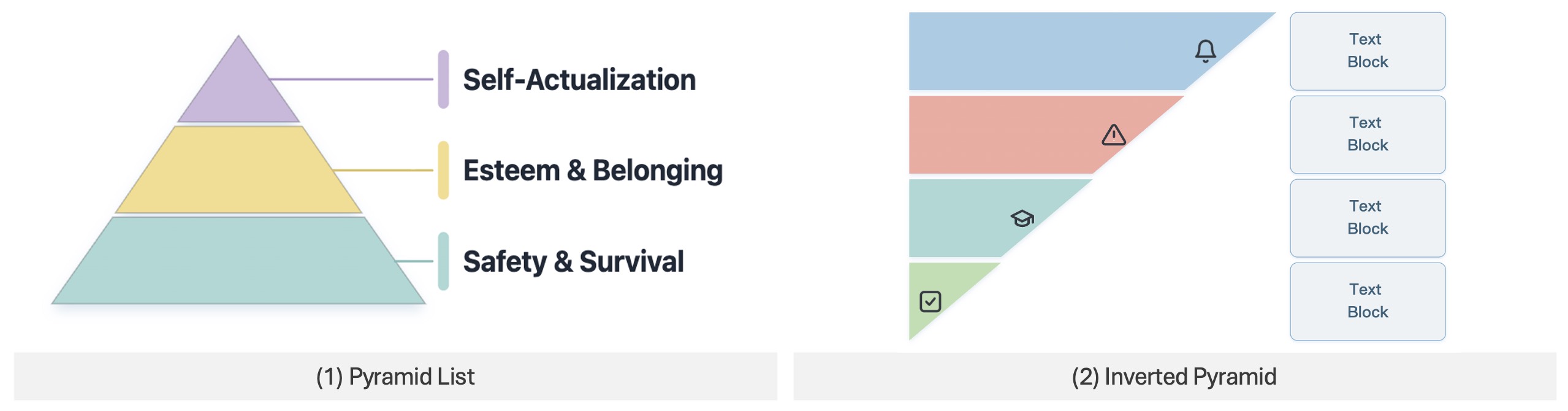} 
    \caption{\textsc{Pyramid}: Pyramid List and Inverted Pyramid.}
    \label{fig:pyramid}
\end{figure*}

\begin{figure*}[!t]
    \centering
    \includegraphics[width=\textwidth]{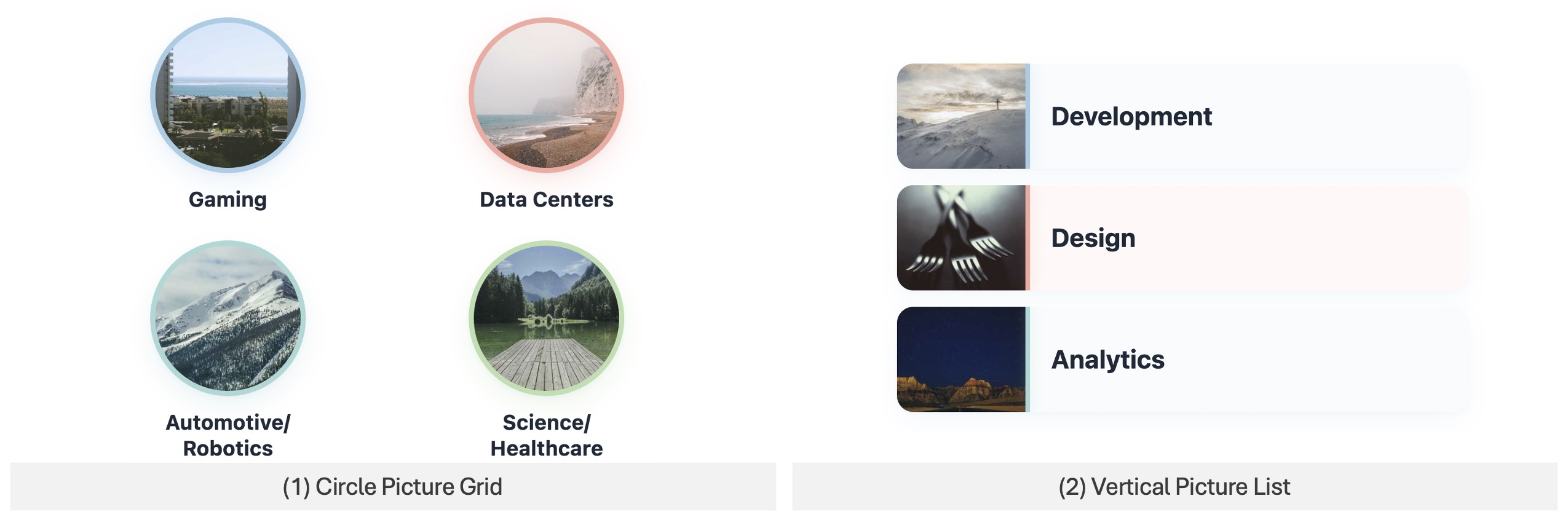} 
    \caption{\textsc{Picture}: Circle Picture Grid and Vertical Picture List.}
    \label{fig:picture}
\end{figure*}

Each template is implemented twice: an HTML version that exposes every node, edge, and container as an addressable DOM element, and a PowerPoint version that realizes the same structure as named vector shapes and connectors. 
Examples from the template library are shown in Figs.~\ref{fig:list}--\ref{fig:picture}, covering the eight logical-relation families from \textsc{List} to \textsc{Picture}.
Importantly, these templates specify only the underlying logical structure; during infographic composition, they are further specialized in visual style, color palette, spacing, and content to match the target infographic design.

\subsubsection{Stage 4: Infographic Composition}
With references and templates in hand, an infographic is composed by 
transferring the color palette of a reference image onto logical-relation 
templates. The HTML track ($800$ infographics) is model-assisted: 
the annotator selects templates from the library that fit the topic and 
logical structure and assembles them; Claude-Opus-4.7 is then invoked in two sequential 
passes, a \emph{styling pass} that adapts the templates' CSS to match the 
reference image's palette, and a \emph{content pass} that 
populates textual placeholders with topic-relevant text from public source, such as Wikipedia. The reference is 
used as a color palette cue, not as content or a layout to reproduce. The resulting 
draft is then post-edited by the annotator to fix layout artifacts, adjust 
visual details, and verify the topic-specific content. 
The PowerPoint track ($200$ infographics) is fully manual: the annotator picks 
templates, arranges them, adjusts colors and typography, and writes the topic 
text. We use these two authoring tracks because HTML and PowerPoint are 
both common and accessible ways of creating infographics in practice, while 
also exposing structured layout objects that can be edited and evaluated 
programmatically. Some templates may include chart-like components to improve visual diversity 
and better reflect real-world infographic design. However, our editing 
instructions focus on infographic-level reflow rather than specific chart-internal 
facts such as numerical values, axes, or plotted trends, whose fine-grained 
factuality has been explored by \citet{zhuo2025factuality} and \citet{li2026charte}.

\looseness=-1
We design the construction pipeline to avoid making the benchmark dependent on the color palette of any single upstream model. Reference images are generated by two different models and are used only to provide high-level cues for color palette. The final infographics are not copied from these references; instead, annotators rebuild them from logical-relation templates and manually post-edit colors, spacing, typography, and layout details. In addition, \ours is an editing benchmark rather than a text-to-image generation benchmark: models are evaluated on modifying an existing infographic, not on generating a new image from the reference. This design makes the benchmark less sensitive to the color preferences of the reference generators.
    
\subsubsection{Stage 5: Quality Control}
    Every infographic is reviewed by a second annotator who did not 
    participate in its creation, against three criteria: 
    (i)~\emph{structural integrity}, every logical-relation instance is complete
and renders without overflow or broken connectors; (ii)~\emph{stylistic faithfulness}, the 
    rendered output matches the reference image in palette; 
(iii)~\emph{content completeness}, all textual fields and labels are coherent, 
topic-relevant, fully readable, and compatible with their logical roles, with no 
placeholders, truncation, duplicated boilerplate, or garbled text.

\subsubsection{Topic Distribution}
\label{appx:topic}

\begin{figure*}[!t]
    \centering
    \includegraphics[width=\textwidth]{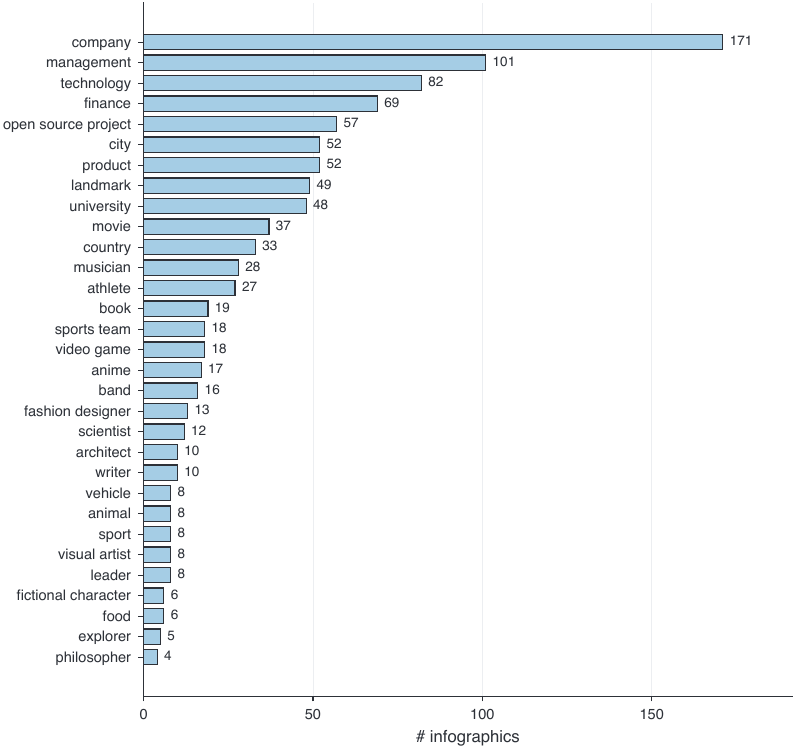}
    \caption{Distribution of infographic topics across 31 categories in \ours, demonstrating broad coverage of real-world domains.}
    \label{fig:topic_stats}
\end{figure*}

The final set of $1{,}000$ infographics covers a broad range of real-world topics. 
As shown in Fig.~\ref{fig:topic_stats}, the distribution spans $31$ topic categories, 
including companies, management, technology, finance, open-source projects, cities, 
products, landmarks, universities, entertainment, sports, and public figures. 
This diversity is intended to reduce topic-specific bias and ensure that \ours evaluates 
structured-content editing across varied infographic use cases rather than a narrow domain.

\subsection{Editing Tasks}
\label{appx:editing_tasks}

\subsubsection{Difficulty-Level Statistics}
\label{appx:difficulty_statistics}

For the three editing tasks with an explicit magnitude axis, we assign each instruction to a difficulty level according to the size of the requested structural perturbation. 
For \textsc{Expand-Text}, difficulty is determined by the number of additional words appended to the target element. 
For \textsc{Insert-Element}, difficulty is determined by the number of new elements inserted into the existing structure. 
For \textsc{Swap-Block}, difficulty is determined by the number of block pairs to be exchanged. 
\textsc{Reshape-Canvas} does not use a discrete magnitude axis; its difficulty depends instead on the source--target aspect-ratio transformation, since different transformations impose different degrees of global re-layout. 
Table~\ref{tab:difficulty_statistics} reports the number of samples at each difficulty level.

\begin{table*}[t]
\centering
\small
\setlength{\tabcolsep}{6pt}
\renewcommand{\arraystretch}{1.12}
\begin{tabular}{@{}llrcc@{}}
\toprule
\textbf{Task} & \textbf{Magnitude Bucket} & \textbf{\#Samples} & \textbf{Ratio} & \textbf{Difficulty} \\
\midrule
\multirow{5}{*}{\textsc{Expand-Text}}
& 5--10 words   & 285 & 28.50\% & Easy \\
& 11--15 words  & 302 & 30.20\% & Medium \\
& 16--20 words  & 205 & 20.50\% & Medium \\
& 21--25 words  &  99 &  9.90\% & Hard \\
& 26--30 words  & 109 & 10.90\% & Hard \\
\midrule
\multirow{5}{*}{\textsc{Insert-Element}}
& 1 element   & 170 & 17.00\% & Easy \\
& 2 elements  & 323 & 32.30\% & Medium \\
& 3 elements  & 257 & 25.70\% & Medium \\
& 4 elements  & 174 & 17.40\% & Hard \\
& 5 elements  &  76 &  7.60\% & Hard \\
\midrule
\multirow{3}{*}{\textsc{Swap-Block}}
& 1 pair   & 432 & 43.20\% & Easy \\
& 2 pairs  & 420 & 42.00\% & Medium \\
& 3 pairs  & 148 & 14.80\% & Hard \\
\bottomrule
\end{tabular}
\caption{Difficulty-level statistics for editing tasks with explicit magnitude axes. Counts are reported over $1{,}000$ instructions per task.}
\label{tab:difficulty_statistics}
\end{table*}

\subsubsection{Addressable Tasks}
\label{appx:addressable_tasks}

All editing instructions in \ours{} are manually annotated to be \emph{addressable}: each requested edit must correspond to a concrete and feasible structural operation on the original infographic. Since every infographic is paired with an editable source file, annotators can verify addressability against the underlying HTML or PowerPoint objects rather than relying only on visual intuition from the rendered image. Annotators verify that the target element or block can be unambiguously identified and that the requested change can be completed without removing content or breaking the underlying logical relation.

Specifically, \textsc{Expand-Text} instructions target text fields that can reasonably support longer content through wrapping, resizing, or local reflow. \textsc{Insert-Element} instructions are applied only to relation instances that can accommodate the requested new elements while preserving their semantics. \textsc{Swap-Block} instructions require both targets to be coherent layout modules that can move as whole units. \textsc{Reshape-Canvas} instructions are kept only when the original content can fit the target aspect ratio through global re-layout rather than content removal.

We further require that the task is non-trivial: the edit should not be solvable by simply pasting content into empty whitespace, but should require some degree of reflow. A second annotator reviews each instruction for target clarity, feasibility, and non-triviality; invalid or underspecified instructions are revised or discarded.

\subsection{Data Examples}
\label{appx:data_examples}
\begin{figure*}[t]
    \centering
    \includegraphics[width=\textwidth]{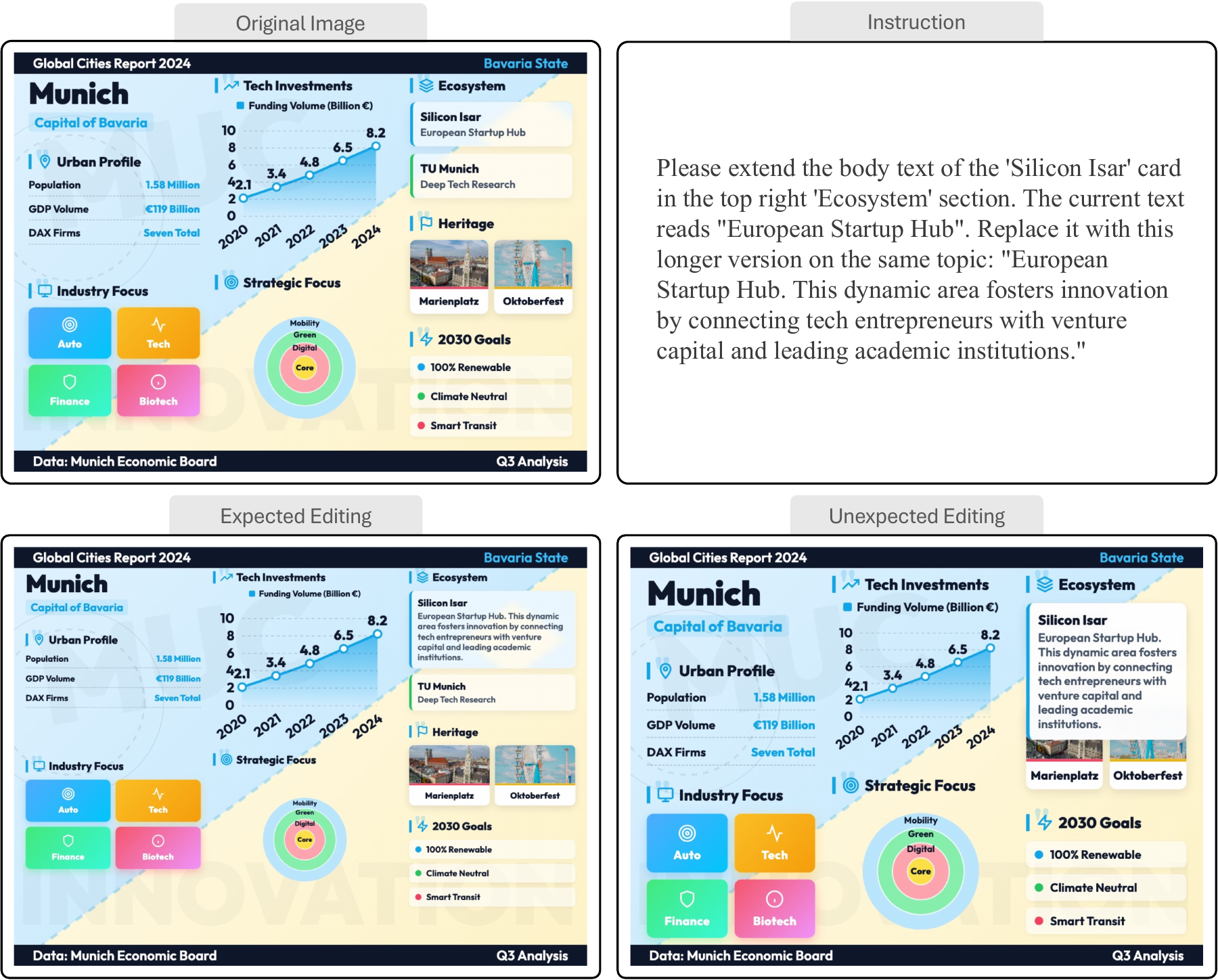}
    \caption{Example of the \textsc{Expand-Text} task in \ours. The editor must lengthen the body text of a specified card while preserving its content. The expected edit reflows nearby elements to maintain readability and layout consistency, whereas the unexpected edit inserts the longer text without properly adjusting the local layout.}
    \label{fig:data_example_expand_text}
\end{figure*}

\begin{figure*}[!t]
    \centering
    \includegraphics[width=\textwidth]{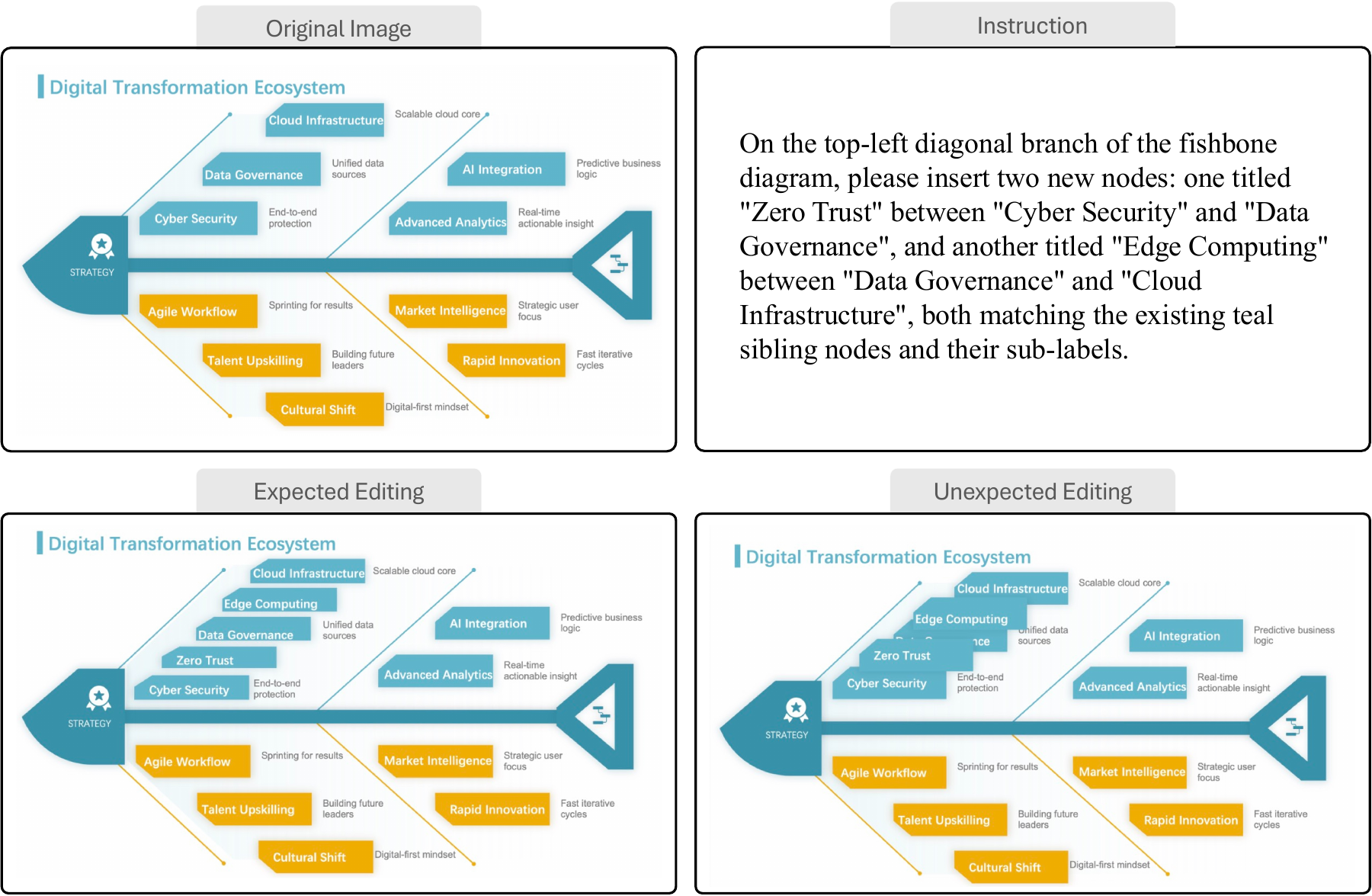}
    \caption{Example of the \textsc{Insert-Element} task in \ours. The editor must insert new nodes into an existing logical structure. The expected edit places them at the specified positions and adjusts the surrounding structure, whereas the unexpected edit adds the nodes without sufficient reflow, resulting in an invalid local layout.}
    \label{fig:data_example_add_element}
\end{figure*}

\begin{figure*}[!t]
    \centering
    \includegraphics[width=\textwidth]{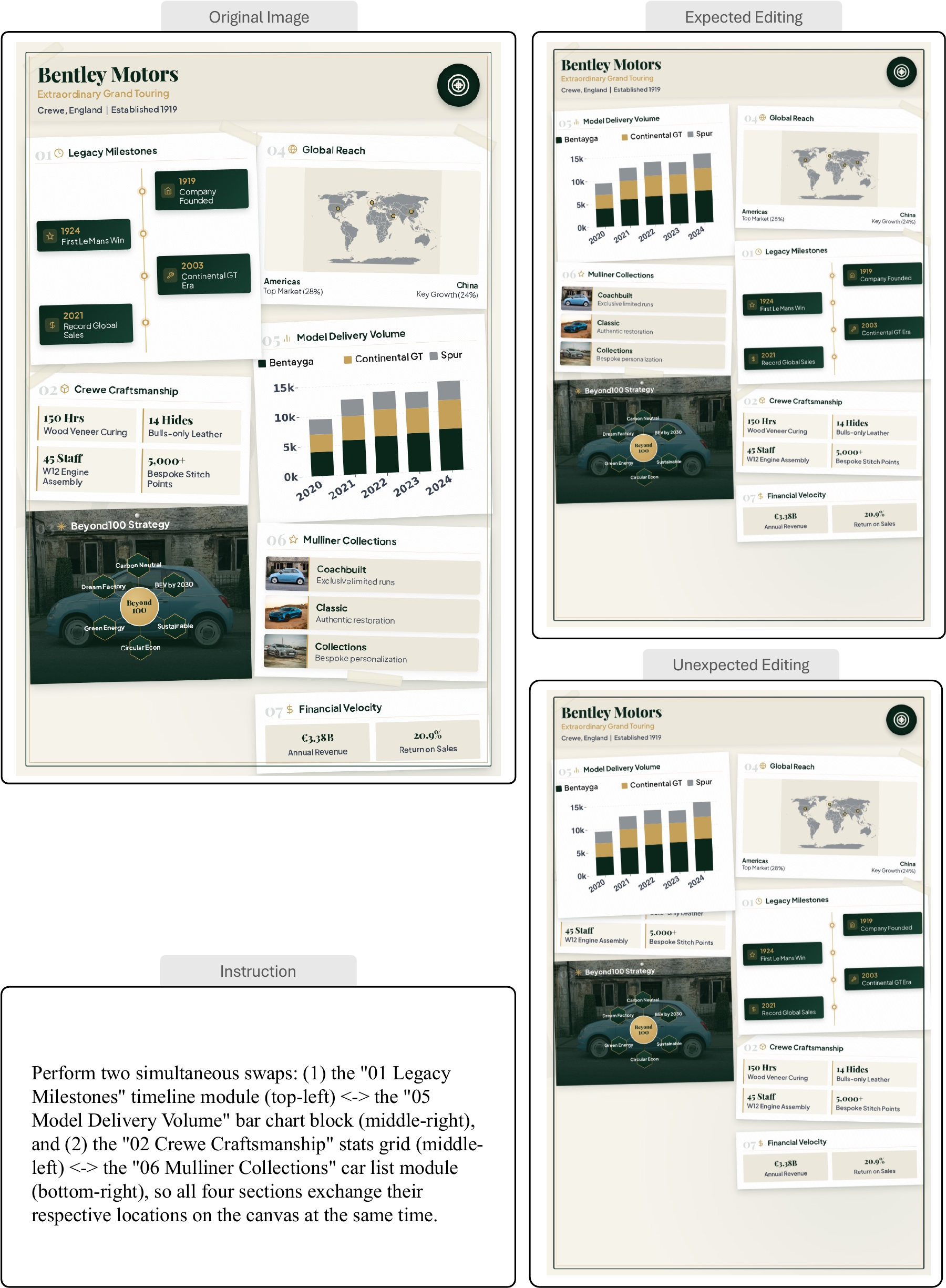}
    \caption{Example of the \textsc{Swap-Block} task in \ours. The instruction asks the editor to exchange multiple logical blocks on the canvas. The expected edit swaps the specified blocks as whole units while preserving their internal content and adapting the surrounding layout; the unexpected edit only partially satisfies the requested swaps or damages the original layout structure.}
    \label{fig:data_example_swap_block}
\end{figure*}

\begin{figure*}[!t]
    \centering
    \includegraphics[width=\textwidth]{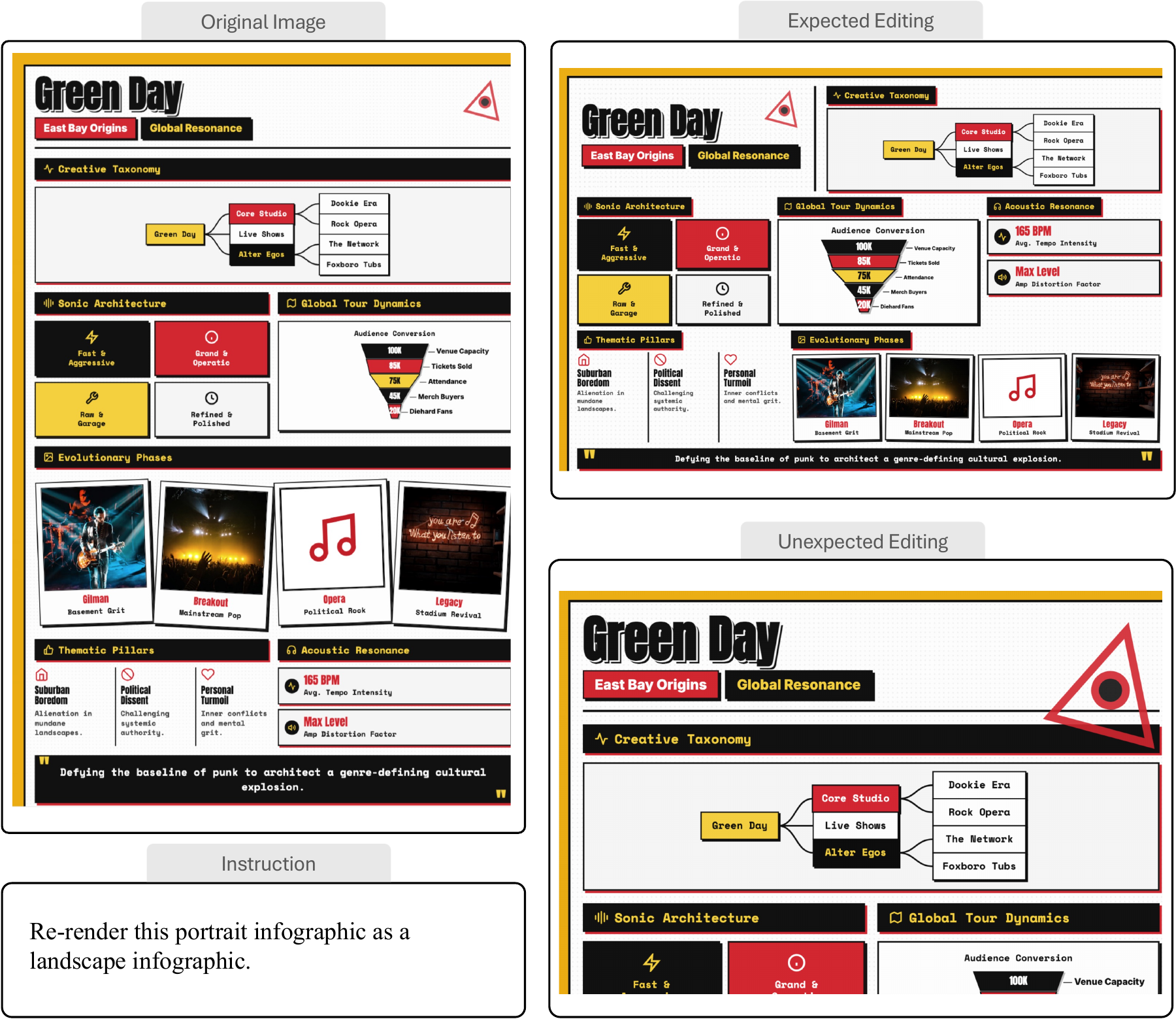}
    \caption{Example of the \textsc{Reshape-Canvas} task in \ours. The instruction asks the editor to re-render a portrait infographic as a landscape infographic. The expected edit globally reflows the content to fit the target aspect ratio while preserving the original information and visual style; the unexpected edit changes the canvas shape but fails to perform a complete global re-layout.}
    \label{fig:data_example_reshape_canvas}
\end{figure*}

We provide representative examples for the four editing tasks in Figs.~\ref{fig:data_example_expand_text}--\ref{fig:data_example_reshape_canvas}.
Each example contains the original infographic, the natural-language editing instruction, an expected edited image, and an unexpected edited image. The expected edited image illustrates a valid edit that satisfies both edit compliance and content preservation, while the unexpected edited image highlights typical failures where the requested change is incomplete or the surrounding structured content is not properly reflowed. Together, these examples illustrate how \ours evaluates not only whether the target edit is executed, but also whether the rest of the infographic remains readable, coherent, and logically valid after reflow.

\section{Additional Related Work}
\label{appx:related_work}

\paragraph{Image Editing Models.}
Instruction-based image editing has progressed from diffusion-specialized 
pipelines~\cite{brooks2023instructpix2pix, sheynin2024emu, hui2024hqedit} 
to unified multimodal models with a shared backbone for image 
understanding and generation~\cite{huang2025diffusion, deng2025bagel, 
wu2025omnigen2, lin2025uniworld}. The current generation, both 
closed-source~\cite{openai2026gptimage2, google2025gemini3image, seedream2026seedream5lite} and 
open-source~\cite{liu2025step1xedit, skywork2025unipic2, zeng2025draw, 
labs2025fluxkontext, wu2025qwenimage}, performs multi-turn, 
identity-preserving edits at near-production quality. Yet across this 
lineage, models are trained and ranked under one implicit assumption: 
the instruction names a region, the model edits that region, and the 
rest of the canvas is held pixel-identical to the source.

\paragraph{Code-Level vs. Pixel-Level Structured Visual Generation and Editing.}
Structured visual content can be generated and edited through two broad pathways. The code-level pathway represents visual artifacts as editable source, such as HTML, PowerPoint, or diagram code, and then renders the source into the final visual output~\cite{jaisankar2024postdoc,mondal2024scidoc2diagrammer, maheshwari2024presentations,ge2025autopresent,zheng2025pptagent,yang2025chartmimic,su2026vcg,huang2025pptbench}. This representation makes structural elements addressable and enables deterministic rendering, but it also requires access to editable source and robust program-level generation or modification. Historically, this code-level pathway has dominated structured visual generation, since earlier image-generation models struggled to render complete and legible text, let alone produce coherent infographics. Recent pixel-level models, however, such as Nano Banana and GPT-Image-2, can render readable text and generate visually appealing infographics, motivating renewed interest in direct pixel-level generation and editing. In contrast, the pixel-level pathway directly produces or edits rendered images, as in recent structured-content generation and editing systems~\cite{openai2026gptimage2,google2025gemini3image,seedream2026seedream5lite,wu2025qwenimage,google2025stitch,NotebookLM,zhu2026paperbanana,shi2026vlms,zhu2026autofigure}. This pathway is more general when source files are unavailable, but it must infer or maintain the latent structure of the visual artifact from pixels alone. Faithfully realizing textual specifications in generated images remains challenging across multimodal generation settings~\cite{yang2026ureason,tao2026asymmetric}.

Which pathway is better suited for structured visual content remains an open question. In Sec.~\ref{sec:discussion}, we take a first step toward examining this question in the editing setting by comparing pixel-level image editors with code-level editing over the editable sources provided in \ours{}. This comparison is not intended as a like-for-like competition, since the code-level pathway has privileged access to editable source objects, while pixel-level editors operate only on rendered images. Instead, our goal is to characterize the trade-off between two practical deployment regimes: source-available editing and source-free image editing. Our results show that the two pathways achieve comparable overall performance but exhibit complementary strengths across tasks. In particular, editable sources provide strong structural addressability for operations such as \textsc{Swap-Block}, while pixel-level editors remain more general when source files are unavailable. These findings suggest that future systems may benefit from combining pixel-level visual flexibility with code-level structural addressability.

\section{Details on Interactive Selection Hint}
\label{app:interactive-selection-hints}

\begin{figure*}[!t]
    \centering
    \includegraphics[width=1.0\linewidth]{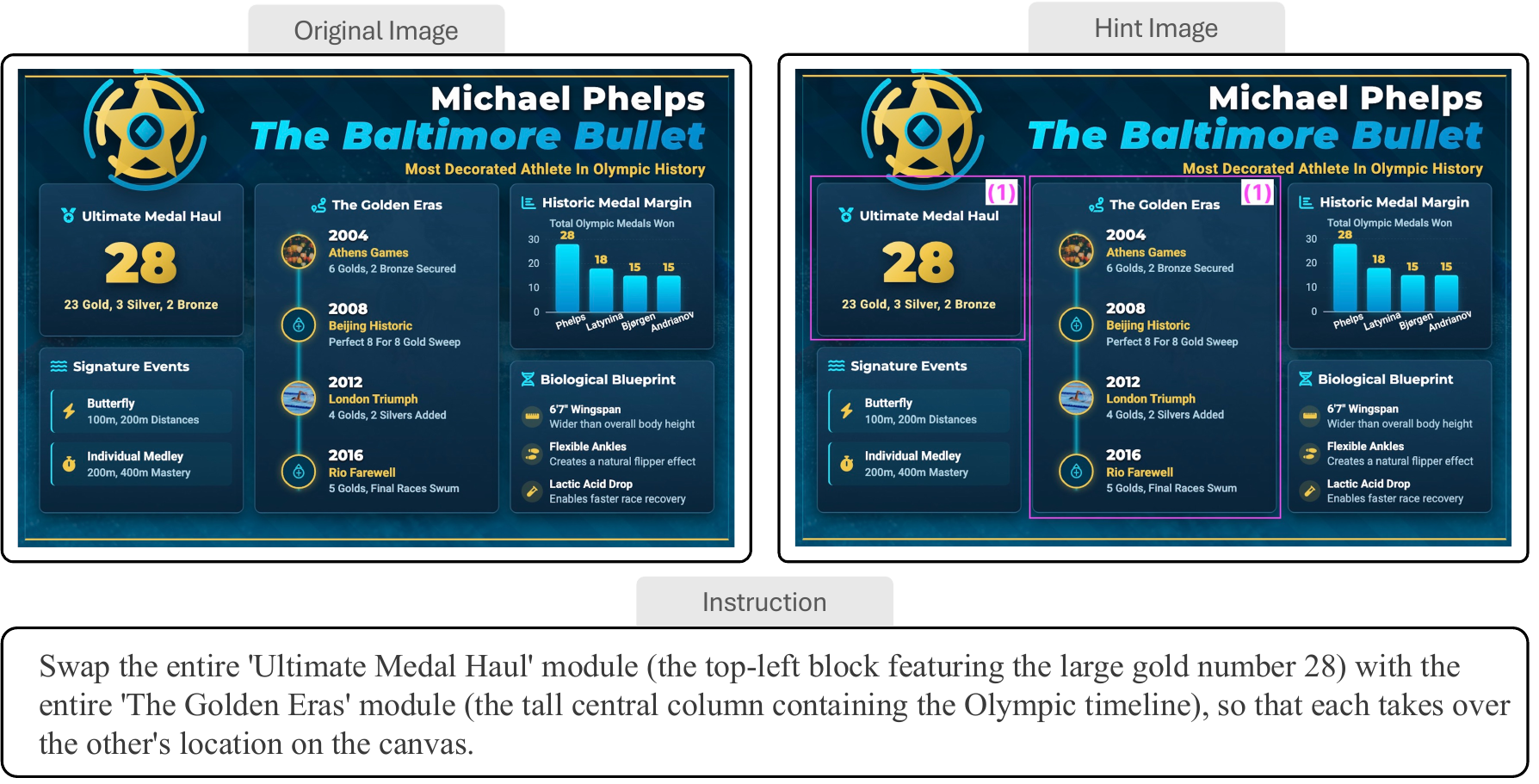}
    \caption{
    Interactive selection hint for \textsc{Swap-Block}. Matching numeric labels indicate the
    two blocks to be exchanged. The boxes are used only for target localization and should not
    appear in the edited output.
    }
    \label{fig:interactive-selection-hint}
\end{figure*}

For the hinted version of \textsc{Swap-Block}, we add visual localization marks to the input image. Each target block is annotated with a tight bounding box, and blocks assigned the same numeric label form a swap pair. For example, two blocks labeled ``(1)'' are swapped. The annotation is made at the block level, where the box covers the whole logical module, including its title, text, icons, charts, connectors, and local container when applicable.

The marked image is used only as an auxiliary input to help the model identify the swap targets.
\looseness=-1
The editing instruction is unchanged except for an additional note that the boxes and labels are selection hints and must not appear in the final output. Fig.~\ref{fig:interactive-selection-hint} shows an example. The two modules highlighted with the same label are the blocks to be exchanged.

\section{Details on Code Generation Baselines}
\label{appx:code_generation}

\looseness=-1
We evaluate code generation as an alternative editing paradigm for 
\ours. Instead of directly modifying the rendered image, a model is 
given the underlying source representation of the infographic and asked to 
apply the same natural-language editing instruction at the code level. The 
modified source is then rendered back to an image, and the resulting PNG is 
evaluated with the same MLLM-as-judge protocol used for image-generation 
editors.

We consider two input settings. In the \emph{code} setting, the model 
receives only the editable source and the instruction. In the 
\emph{code\,+\,image} setting, the model additionally receives the original 
rendered infographic as visual grounding, so that it can better infer the 
visual layout, styling, and spatial relationships while still producing code as 
output.

\looseness=-1
The implementation differs slightly across the two source formats. For 
HTML infographics, the model rewrites the standalone HTML document, which 
is subsequently rendered to PNG using a browser. For PowerPoint infographics, 
we first expose a structured textual dump of the slide, including shapes, text, 
positions, and sizes; the model then produces a Python editing function using 
\texttt{python-pptx}, and the edited slide is rendered to PNG. 
\begingroup
\section{Evaluation Reliability}
\label{appx:human_evaluation}

\subsection{Human Agreement across Output Models}

We validate the binary overall verdict against expert human judgments across
three output models spanning different capability levels. For each model, we
use $800$ outputs under the same protocol, with exactly $200$ examples from
each editing task. Three expert annotators independently assign a binary
success or failure label, and their majority label is treated as the human
reference. Table~\ref{tab:human_agreement_models} reports agreement between
Gemini-3.1-Pro and the human majority. The high agreement on
HunyuanImage-3.0 should be interpreted together with its output distribution:
most outputs are clear failures, which are straightforward for both the human
annotators and the judge to identify.

\begin{table}[h]
\centering
\small
\setlength{\tabcolsep}{7pt}
\begin{tabular}{lrr}
\toprule
\textbf{Output model} & \textbf{Samples} & \textbf{Agreement} \\
\midrule
GPT-Image-2       & 800 & 0.920 \\
Nanobanana-2      & 800 & 0.934 \\
HunyuanImage-3.0  & 800 & 0.981 \\
\bottomrule
\end{tabular}
\caption{Agreement between Gemini-3.1-Pro and the human-majority verdict
across output models.}
\label{tab:human_agreement_models}
\end{table}

\subsection{Stratified Agreement}

For GPT-Image-2, inter-annotator agreement is $0.940$. We additionally
stratify judge--human agreement by task, requested-edit difficulty, and the
judge's predicted outcome. As shown in
Table~\ref{tab:human_agreement_strata}, agreement remains high throughout.
The difficulty analysis covers the three tasks with explicit magnitude axes;
\textsc{Reshape-Canvas} is excluded because it has no easy/medium/hard
magnitude label. The outcome analysis is a post-hoc split by the judge's
binary verdict.

\begin{table}[h]
\centering
\small
\begin{tabular}{llr}
\toprule
\textbf{Axis} & \textbf{Subset} & \textbf{Agreement} \\
\midrule
\multirow{4}{*}{Task}
 & \textsc{Expand-Text}    & 0.920 \\
 & \textsc{Insert-Element} & 0.905 \\
 & \textsc{Swap-Block}     & 0.925 \\
 & \textsc{Reshape-Canvas} & 0.930 \\
\midrule
\multirow{3}{*}{Difficulty}
 & Easy   & 0.943 \\
 & Medium & 0.912 \\
 & Hard   & 0.890 \\
\midrule
\multirow{2}{*}{Judge verdict}
 & Success & 0.913 \\
 & Failure & 0.931 \\
\bottomrule
\end{tabular}
\caption{Gemini-3.1-Pro agreement with the human majority on GPT-Image-2
outputs, stratified by task, difficulty, and predicted outcome.}
\label{tab:human_agreement_strata}
\end{table}
\endgroup

\begingroup
\subsection{Alternative Judge}
\label{appx:alternative_evaluators}

We test whether evaluation reliability depends on the choice of MLLM judge by
re-evaluating the same model outputs with Claude-Opus-4.7 under the same rubric
and prompt. Table~\ref{tab:alternative_judges} compares both judges against
the human-majority labels. The two cross-vendor judges obtain similar
agreement on outputs from both models.

\begin{table}[h]
\centering
\small
\begin{tabular}{lcc}
\toprule
\textbf{Judge} & \textbf{GPT-Image-2} & \textbf{Nanobanana-2} \\
\midrule
Gemini-3.1-Pro  & 0.920 & 0.934 \\
Claude-Opus-4.7 & 0.914 & 0.928 \\
\bottomrule
\end{tabular}
\caption{Agreement with the human-majority verdict for two MLLM judges.}
\label{tab:alternative_judges}
\end{table}

\subsection{Repeated Evaluation}

We also run Gemini-3.1-Pro five times on the same $800$ outputs per model.
Table~\ref{tab:repeated_judge} reports the mean agreement and standard
deviation across runs. The small variation indicates that evaluation
randomness does not affect the reported comparison.

\begin{table}[h]
\centering
\small
\begin{tabular}{lc}
\toprule
\textbf{Output model} & \textbf{Agreement over five runs} \\
\midrule
GPT-Image-2  & $0.924 \pm 0.007$ \\
Nanobanana-2 & $0.936 \pm 0.008$ \\
\bottomrule
\end{tabular}
\caption{Repeated Gemini-3.1-Pro evaluation against human-majority labels.}
\label{tab:repeated_judge}
\end{table}
\endgroup

\begingroup
\section{Fine-Grained Results}
\label{appx:fine_grained}

\subsection{Logical-Relation Families}

We report per-family results on \textsc{Insert-Element}, because every
instruction in this task targets a specific logical-relation block and can be
cleanly attributed to its host family. Table~\ref{tab:relation_breakdown}
shows that GPT-Image-2 leads across all families. Cycle is the hardest family
for all three editors, consistent with the need to reroute and reclose a loop
of connectors after insertion.

\begin{table}[h]
\centering
\small
\resizebox{\columnwidth}{!}{%
\begin{tabular}{lrrr}
\toprule
\textbf{Family} & \textbf{GPT-Image-2} & \textbf{Nanobanana-Pro} & \textbf{Nanobanana-2} \\
\midrule
List         & 78.9 & 45.3 & 40.5 \\
Matrix       & 75.2 & 37.0 & 34.5 \\
Process      & 77.1 & 47.3 & 45.0 \\
Picture      & 75.5 & 36.5 & 35.2 \\
Cycle        & 49.3 & 20.6 & 31.6 \\
Pyramid      & 72.7 & 49.5 & 46.5 \\
Relationship & 67.1 & 38.4 & 39.7 \\
Hierarchy    & 78.7 & 63.8 & 48.9 \\
\bottomrule
\end{tabular}%
}
\caption{\textsc{Insert-Element} SR by host logical-relation family. All
values are percentages.}
\label{tab:relation_breakdown}
\end{table}

\subsection{HTML versus PowerPoint Sources}

Table~\ref{tab:source_format_pixel} reports pixel-level editing results by
the source format used to author the input infographic. Because pixel-level
editors receive only the rendered image, their results are similar between
HTML and PowerPoint, and their ranking is unchanged.

\begin{table*}[t]
\centering
\scriptsize
\setlength{\tabcolsep}{3.5pt}
\resizebox{\textwidth}{!}{%
\begin{tabular}{ll rrr rrr rrr rrr r}
\toprule
& & \multicolumn{3}{c}{\textbf{\textsc{Expand-Text}}}
& \multicolumn{3}{c}{\textbf{\textsc{Insert-Element}}}
& \multicolumn{3}{c}{\textbf{\textsc{Swap-Block}}}
& \multicolumn{3}{c}{\textbf{\textsc{Reshape-Canvas}}}
& \textbf{Avg.} \\
\cmidrule(lr){3-5}\cmidrule(lr){6-8}\cmidrule(lr){9-11}\cmidrule(lr){12-14}
\textbf{Model} & \textbf{Source} & EC & CP & SR & EC & CP & SR & EC & CP & SR & EC & CP & SR & SR \\
\midrule
\multirow{3}{*}{GPT-Image-2}
 & Overall & 99.0 & 86.7 & 85.9 & 85.4 & 80.7 & 72.0 & 29.0 & 79.2 & 25.5 & 96.3 & 65.1 & 64.4 & 62.0 \\
 & HTML    & 99.1 & 86.3 & 85.5 & 85.8 & 80.3 & 71.9 & 29.5 & 78.9 & 25.9 & 96.6 & 64.6 & 64.0 & 61.8 \\
 & PPT     & 98.5 & 88.5 & 87.5 & 84.0 & 82.5 & 72.5 & 27.0 & 80.5 & 24.0 & 95.0 & 67.0 & 66.0 & 62.5 \\
\midrule
\multirow{3}{*}{Nanobanana-2}
 & Overall & 85.1 & 58.6 & 52.6 & 65.3 & 49.4 & 39.0 & 30.0 & 72.1 & 24.0 & 92.0 & 36.4 & 36.3 & 38.0 \\
 & HTML    & 84.9 & 58.1 & 52.1 & 65.9 & 48.9 & 38.9 & 30.5 & 71.8 & 24.4 & 92.3 & 35.9 & 35.9 & 37.8 \\
 & PPT     & 86.0 & 60.5 & 54.5 & 63.0 & 51.5 & 39.5 & 28.0 & 73.5 & 22.5 & 91.0 & 38.5 & 38.0 & 38.6 \\
\bottomrule
\end{tabular}%
}
\caption{Pixel-level editing results by source format.}
\label{tab:source_format_pixel}
\end{table*}

Table~\ref{tab:source_format_code} gives the corresponding breakdown for
code-level editing with both source code and rendered image as input. In this
setting, both models perform substantially better on HTML than PowerPoint.

\begin{table*}[t]
\centering
\scriptsize
\setlength{\tabcolsep}{3.5pt}
\resizebox{\textwidth}{!}{%
\begin{tabular}{ll rrr rrr rrr rrr r}
\toprule
& & \multicolumn{3}{c}{\textbf{\textsc{Expand-Text}}}
& \multicolumn{3}{c}{\textbf{\textsc{Insert-Element}}}
& \multicolumn{3}{c}{\textbf{\textsc{Swap-Block}}}
& \multicolumn{3}{c}{\textbf{\textsc{Reshape-Canvas}}}
& \textbf{Avg.} \\
\cmidrule(lr){3-5}\cmidrule(lr){6-8}\cmidrule(lr){9-11}\cmidrule(lr){12-14}
\textbf{Model} & \textbf{Source} & EC & CP & SR & EC & CP & SR & EC & CP & SR & EC & CP & SR & SR \\
\midrule
\multirow{3}{*}{Gemini-3.1-Pro}
 & Overall & 88.2 & 72.0 & 68.5 & 85.3 & 72.2 & 71.0 & 69.0 & 78.4 & 60.2 & 71.8 & 47.2 & 46.8 & 61.6 \\
 & HTML    & 96.5 & 79.6 & 77.6 & 92.6 & 80.9 & 79.5 & 75.6 & 85.4 & 68.0 & 75.9 & 53.4 & 52.9 & 69.5 \\
 & PPT     & 55.0 & 41.5 & 32.0 & 56.0 & 37.5 & 37.0 & 42.5 & 50.5 & 29.0 & 55.5 & 22.5 & 22.5 & 30.1 \\
\midrule
\multirow{3}{*}{Gemini-3.5-Flash}
 & Overall & 85.1 & 69.5 & 65.8 & 85.4 & 67.2 & 63.2 & 60.9 & 79.8 & 56.8 & 71.2 & 46.6 & 43.1 & 57.2 \\
 & HTML    & 93.1 & 76.9 & 74.5 & 92.8 & 75.6 & 70.8 & 66.8 & 86.9 & 64.1 & 75.1 & 52.8 & 48.8 & 64.5 \\
 & PPT     & 53.0 & 40.0 & 31.0 & 56.0 & 33.5 & 33.0 & 37.5 & 51.5 & 27.5 & 55.5 & 22.0 & 20.5 & 28.0 \\
\bottomrule
\end{tabular}%
}
\caption{Code-level editing with code and image input, broken down by source format.}
\label{tab:source_format_code}
\end{table*}

\subsection{Canvas Transformation Types}

We categorize \textsc{Reshape-Canvas} instructions by their source-to-target
aspect-ratio direction. Table~\ref{tab:reshape_directions} shows that all
three models are weakest on landscape-to-portrait transformations.

\begin{table*}[!h]
\centering
\small
\resizebox{\columnwidth}{!}{%
\begin{tabular}{lrrrr}
\toprule
\textbf{Model} & \textbf{P$\rightarrow$L} & \textbf{L$\rightarrow$P} & \textbf{S$\rightarrow$L} & \textbf{S$\rightarrow$P} \\
\midrule
GPT-Image-2      & 69.1 & 59.4 & 72.5 & 67.5 \\
Nanobanana-Pro   & 36.7 & 31.8 & 65.0 & 60.0 \\
Nanobanana-2     & 41.2 & 32.0 & 35.0 & 37.5 \\
\bottomrule
\end{tabular}%
}
\caption{\textsc{Reshape-Canvas} SR by transformation direction. P, L, and
S denote portrait, landscape, and square.}
\label{tab:reshape_directions}
\end{table*}
\endgroup

\begingroup
\section{Additional Robustness Analyses}
\label{appx:robustness}

\subsection{Color-Palette Sensitivity}

The generated reference images in our construction pipeline are used only as
color-palette cues. To test whether this choice systematically affects editor
performance, we replace each infographic's palette while keeping its layout,
structure, and content unchanged. The replacement palettes are suggested and
applied by GLM-5.2, and annotators verify that only the color scheme changes.
As shown in Table~\ref{tab:color_control}, the two evaluated models move in
opposite directions by small amounts in average SR, and their ranking remains
unchanged.

\begin{table*}[!t]
\centering
\scriptsize
\setlength{\tabcolsep}{3.5pt}
\resizebox{\textwidth}{!}{%
\begin{tabular}{ll rrr rrr rrr rrr r}
\toprule
& & \multicolumn{3}{c}{\textbf{\textsc{Expand-Text}}}
& \multicolumn{3}{c}{\textbf{\textsc{Insert-Element}}}
& \multicolumn{3}{c}{\textbf{\textsc{Swap-Block}}}
& \multicolumn{3}{c}{\textbf{\textsc{Reshape-Canvas}}}
& \textbf{Avg.} \\
\cmidrule(lr){3-5}\cmidrule(lr){6-8}\cmidrule(lr){9-11}\cmidrule(lr){12-14}
\textbf{Model} & \textbf{Setting} & EC & CP & SR & EC & CP & SR & EC & CP & SR & EC & CP & SR & SR \\
\midrule
\multirow{2}{*}{GPT-Image-2}
 & Original     & 99.0 & 86.7 & 85.9 & 85.4 & 80.7 & 72.0 & 29.0 & 79.2 & 25.5 & 96.3 & 65.1 & 64.4 & 62.0 \\
 & Color-Change & 98.2 & 86.2 & 84.8 & 84.2 & 81.3 & 74.2 & 31.0 & 81.2 & 27.2 & 98.2 & 67.1 & 66.8 & 63.3 \\
\midrule
\multirow{2}{*}{Nanobanana-2}
 & Original     & 85.1 & 58.6 & 52.6 & 65.3 & 49.4 & 39.0 & 30.0 & 72.1 & 24.0 & 92.0 & 36.4 & 36.3 & 38.0 \\
 & Color-Change & 86.2 & 57.2 & 51.7 & 63.2 & 49.0 & 37.2 & 32.1 & 73.2 & 24.8 & 92.3 & 35.2 & 34.9 & 37.2 \\
\bottomrule
\end{tabular}%
}
\caption{Color-palette control. All values are percentages.}
\label{tab:color_control}
\end{table*}

\subsection{Multilingual Pilot}

To examine whether the editable-source construction can support multilingual
evaluation, we construct a preliminary Chinese version of the samples. We use
GLM-5.2 to translate the infographic text directly in the editable source,
re-render the sources, and translate the corresponding editing instructions.
Annotators verify that the layout, structure, and logical relations remain
unchanged. Table~\ref{tab:chinese_pilot} shows lower average SR for both
models, while their ranking remains unchanged. This rapid pilot is intended to
demonstrate extensibility rather than serve as a complete multilingual
evaluation.

\begin{table*}[!t]
\centering
\scriptsize
\setlength{\tabcolsep}{3.5pt}
\resizebox{\textwidth}{!}{%
\begin{tabular}{ll rrr rrr rrr rrr r}
\toprule
& & \multicolumn{3}{c}{\textbf{\textsc{Expand-Text}}}
& \multicolumn{3}{c}{\textbf{\textsc{Insert-Element}}}
& \multicolumn{3}{c}{\textbf{\textsc{Swap-Block}}}
& \multicolumn{3}{c}{\textbf{\textsc{Reshape-Canvas}}}
& \textbf{Avg.} \\
\cmidrule(lr){3-5}\cmidrule(lr){6-8}\cmidrule(lr){9-11}\cmidrule(lr){12-14}
\textbf{Model} & \textbf{Setting} & EC & CP & SR & EC & CP & SR & EC & CP & SR & EC & CP & SR & SR \\
\midrule
\multirow{2}{*}{GPT-Image-2}
 & Original & 99.0 & 86.7 & 85.9 & 85.4 & 80.7 & 72.0 & 29.0 & 79.2 & 25.5 & 96.3 & 65.1 & 64.4 & 62.0 \\
 & Chinese  & 93.0 & 84.5 & 82.1 & 80.4 & 74.2 & 67.2 & 25.2 & 75.2 & 24.1 & 93.2 & 61.2 & 57.2 & 57.7 \\
\midrule
\multirow{2}{*}{Nanobanana-2}
 & Original & 85.1 & 58.6 & 52.6 & 65.3 & 49.4 & 39.0 & 30.0 & 72.1 & 24.0 & 92.0 & 36.4 & 36.3 & 38.0 \\
 & Chinese  & 86.2 & 62.1 & 56.4 & 42.5 & 46.2 & 25.3 & 22.6 & 67.2 & 15.2 & 86.2 & 28.2 & 26.3 & 30.8 \\
\bottomrule
\end{tabular}%
}
\caption{Preliminary Chinese translation pilot. All values are percentages.}
\label{tab:chinese_pilot}
\end{table*}
\endgroup

\section{Ethical Statement}

\ours{} is constructed for research purposes to evaluate infographic editing and reflow capabilities. The benchmark does not directly republish existing copyrighted infographic datasets as editable assets. Some samples mention real-world entities such as companies, products, universities or cities. We use only publicly available information, such as Wikipedia or official public pages, and manually review the content to avoid private, sensitive, defamatory, or misleading claims.

\section{Reproducibility}
\label{appx:reproducibility}

Upon publication, we will release \ours{} for research use. For reproducibility, Tab.~\ref{tab:models} reports the exact model identifiers used in our experiments. We will release the code to reproduce the results. Following prior analyses of LLM decoding sensitivity~\cite{shi-etal-2024-thorough}, we set the temperature to $0.1$ for all automated MLLM and LLM evaluators.

\section{Models}
\label{appx:models}
Tab.~\ref{tab:models} summarizes the models used in our experiments and the corresponding inference interfaces. 
As discussed in Appx.~\ref{appx:related_work}, many earlier image-editing models cannot reliably render complete and legible text, making them unsuitable for infographic editing where text is a core component. 
We therefore exclude such models and focus on recent baselines that meet this basic requirement. 
For proprietary models, we report the exact API model identifiers used in our calls; for open-weight models, we report the Hugging Face repositories from which the checkpoints are obtained.

\begin{table*}[h]
\centering
\scriptsize
\begin{tabularx}{\textwidth}{@{}ll>{\raggedright\arraybackslash}X@{}} 
\toprule
\textbf{Model Name} & \textbf{Type} & \textbf{Inference Interface} \\
\midrule
\multicolumn{3}{@{}l}{\textit{Proprietary models}} \\
GPT-Image-2 & API & \texttt{gpt-image-2}  \\
GPT-Image-1.5 & API & \texttt{gpt-image-1.5}  \\
Nanobanana-Pro & API & \texttt{gemini-3-pro-image-preview} \\
Nanobanana-2 & API & \texttt{gemini-3.1-flash-image-preview} \\
Nanobanana & API & \texttt{gemini-2.5-flash-image} \\
Seedream-5.0-Lite & API & \texttt{doubao-seedream-5-0-260128} \\
Gemini-3.1-Pro & API & \texttt{gemini-3.1-pro-preview} \\
Gemini-3.5-Flash & API & \texttt{gemini-3.5-flash} \\
Gemini-3.1-Flash & API & \texttt{gemini-3.1-flash-lite-preview} \\
Claude-Opus-4.7 & API & \texttt{claude-opus-4-7} \\
GPT-5.4 & API & \texttt{gpt-5.4-2026-03-05} \\
\midrule
\multicolumn{3}{@{}l}{\textit{Open-weight models}} \\
HunyuanImage-3.0-Instruct & Hugging Face & \url{https://huggingface.co/tencent/HunyuanImage-3.0-Instruct} \\
Qwen-Image-Edit & Hugging Face & \url{https://huggingface.co/Qwen/Qwen-Image-Edit-2511} \\
FLUX.2-klein-base-9B & Hugging Face & \url{https://huggingface.co/black-forest-labs/FLUX.2-klein-base-9B} \\
\bottomrule
\end{tabularx}
\caption{Models used in our experiments and their inference interfaces. For proprietary models, we report the API model identifier used in our calls; for open-weight models, we report the corresponding Hugging Face repository.}
\label{tab:models}
\end{table*}

\section{Error Analysis}
\label{appx:case_study}

We conducted an error mode analysis on GPT-Image-2, the best-performing model in our evaluation. 
As shown in Fig.~\ref{fig:error_analysis}, we present the distribution of different failure modes per task, including Incomplete Content, Misplacement, Element Loss, Occlusion, Text Rewritten, Style Change, Structural Break, Content Altered, and No Re-layout. We define each error mode as follows:

\begin{itemize}
    \item \textbf{Incomplete Content}: The target element of an add or text-expand operation is not fully rendered. Its content, such as icons, labels, numbers, or expanded body text, is truncated, clipped, or overflows the container.
    \item \textbf{Misplacement}: The target element is misplaced. For Insert-Element, the new element does not reference the specified relative anchor. For Swap-Block, the two blocks are not correctly exchanged.
    \item \textbf{Element Loss}: One or more original elements no longer fully exist on the canvas. They are either deleted outright, pushed off the visible area, or cropped at the edges during reflow.
    \item \textbf{Occlusion}: An original element is partially or fully covered by another element due to incorrect ordering or layout collisions.
    \item \textbf{Text Rewritten}: The text of non-target elements are altered or rewritten.
    \item \textbf{Style Change}: The visual styling of original elements, such as icons, logos, or color palette, is redesigned rather than preserved.
    \item \textbf{Structural Break}: Structural relationships among elements in the canvas are broken, manifesting as redirected connectors in process flows, broken closure in cyclic diagrams, or skipped indices in numbered sequences.
    \item \textbf{Content Altered}: The internal content (titles, icons, text, numbers, background, or visual style) of the swapped blocks is modified during editing.
    \item  \textbf{No Re-layout}: The model fails to genuinely re-arrange the infographic for the new canvas shape, instead treating it as a flat image.
\end{itemize}

\begin{figure*}[t]
    \centering
    \includegraphics[width=\textwidth]{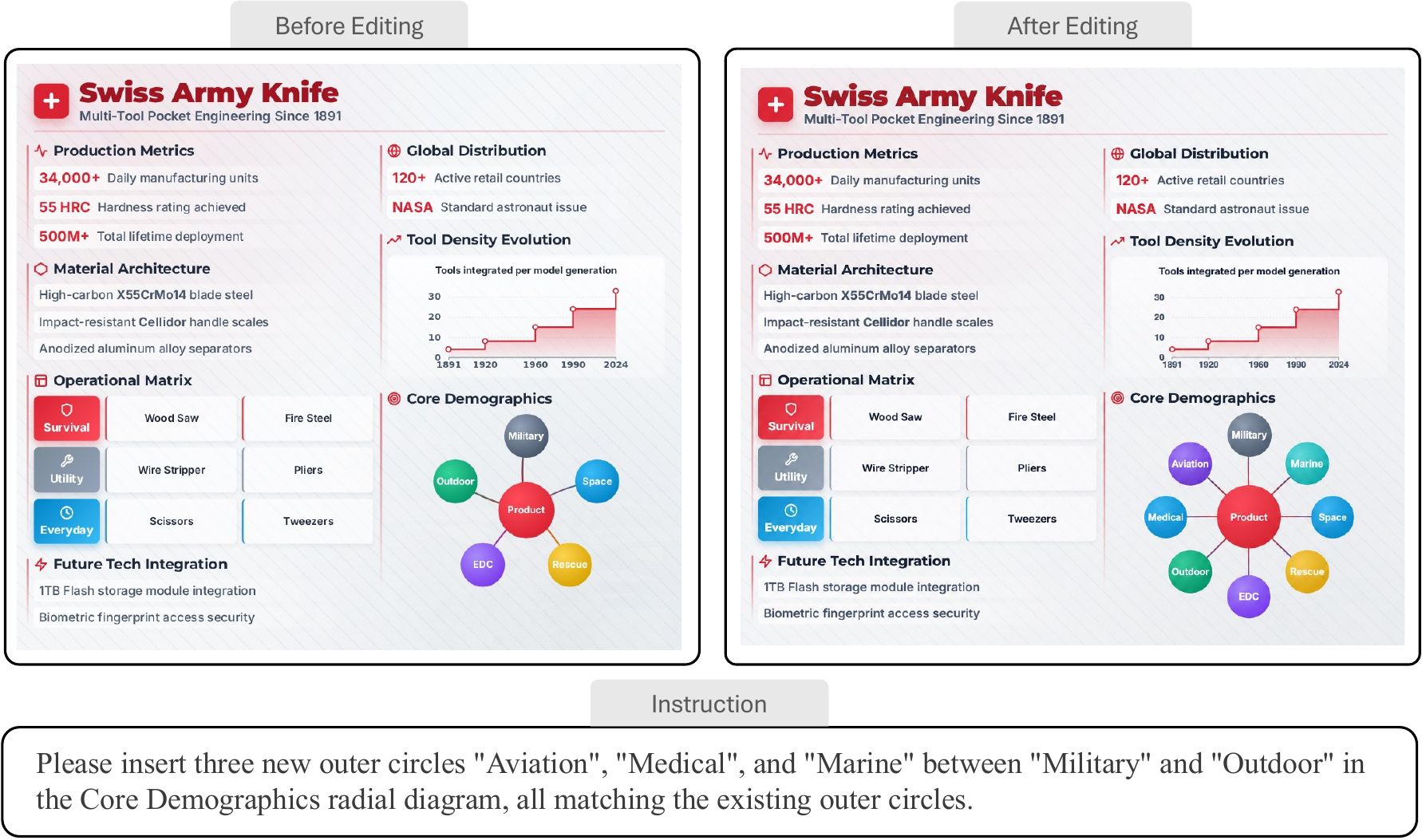} 
    \caption{Misplacement after Insert-Element.
    The three new circles were inserted into the Core Demographics radial diagram, but ``Marine'' is placed in the wrong position. It is incorrectly positioned on the opposite side of the diagram between ``Military'' and ``Space''.}
    \label{fig:bad_case_add_misplacement}
\end{figure*}

\begin{figure*}[!t]
    \centering
    \includegraphics[width=\textwidth]{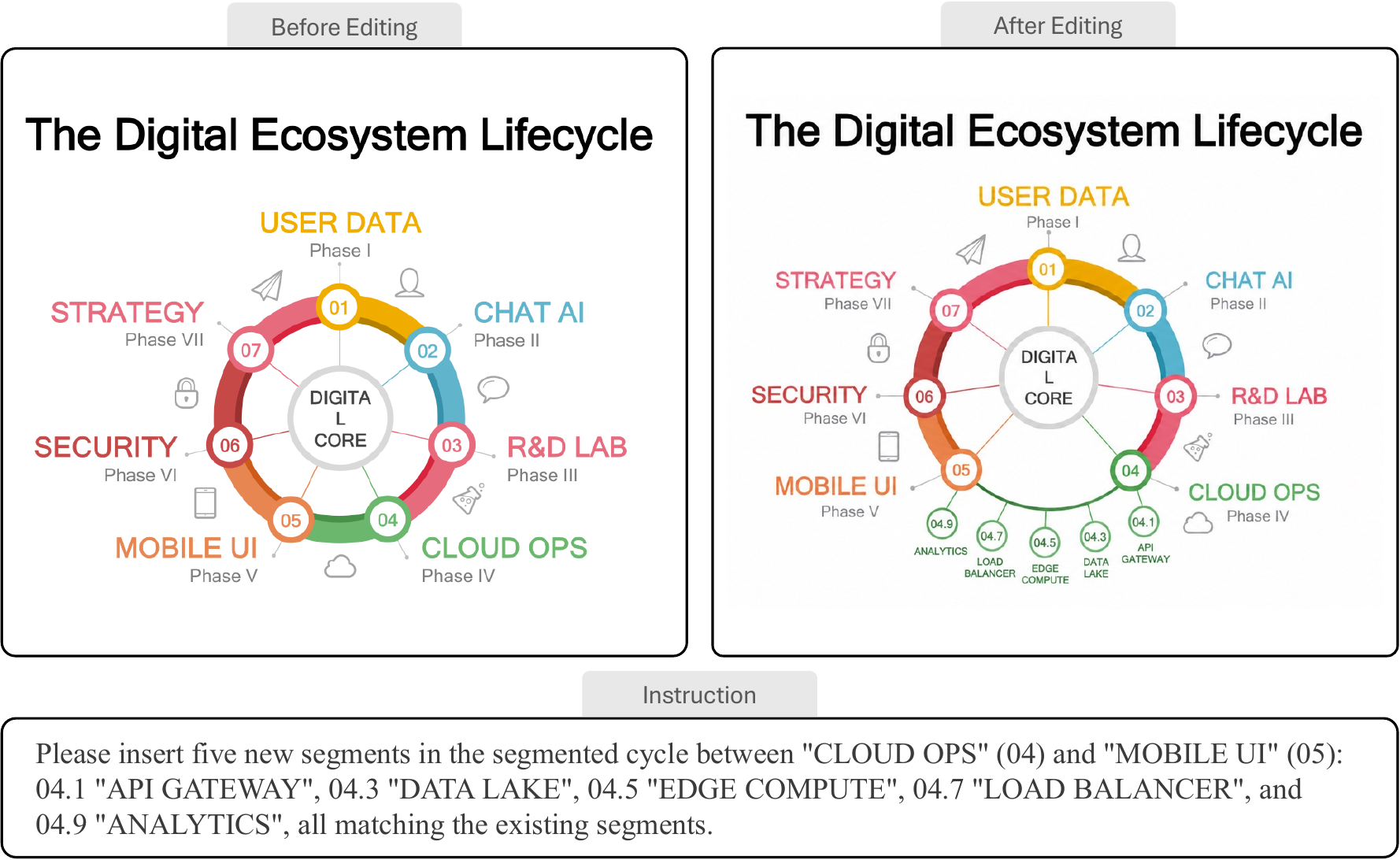} 
    \caption{Style change after Insert-Element.
    The five new segments were inserted between ``Cloud Ops'' and ``Mobile UI'', but they do not match the style of the existing segments. Instead of being integrated as arc segments within the main ring, they are rendered as smaller circular nodes branching outward below the cycle.}
    \label{fig:bad_case_add_style_drift}
\end{figure*}

\begin{figure*}[!t]
    \centering
    \includegraphics[width=\textwidth]{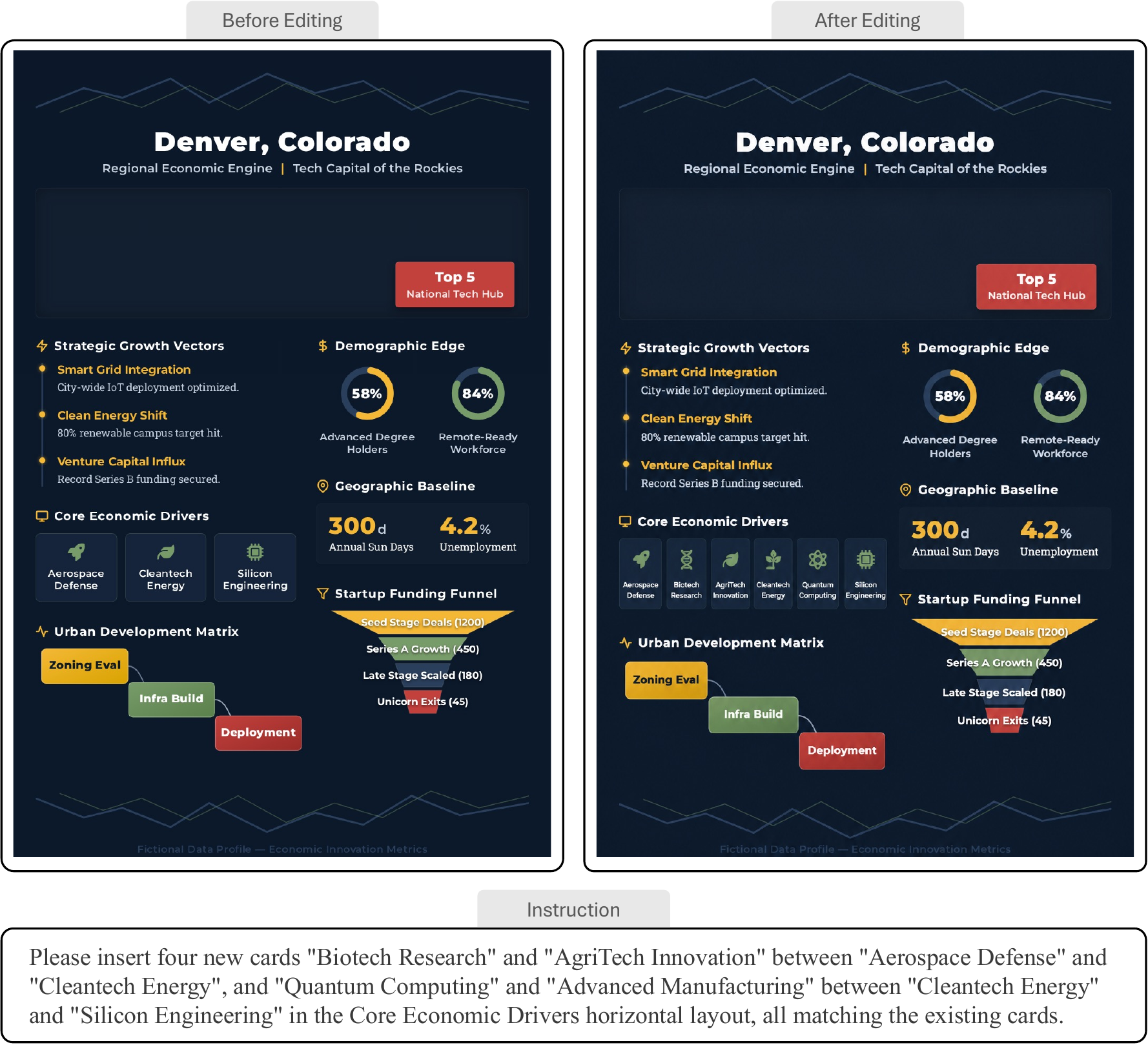} 
    \caption{Incomplete content after Insert-Element. The Core Economic Drivers row is incomplete. There are only three of the four requested cards were inserted, while ``Advanced Manufacturing'' is missing between ``Cleantech Energy'' and ``Silicon Engineering''.}
    \label{fig:bad_case_add_incomplete_content}
\end{figure*}

\begin{figure*}[!t]
    \centering
    \includegraphics[width=\textwidth]{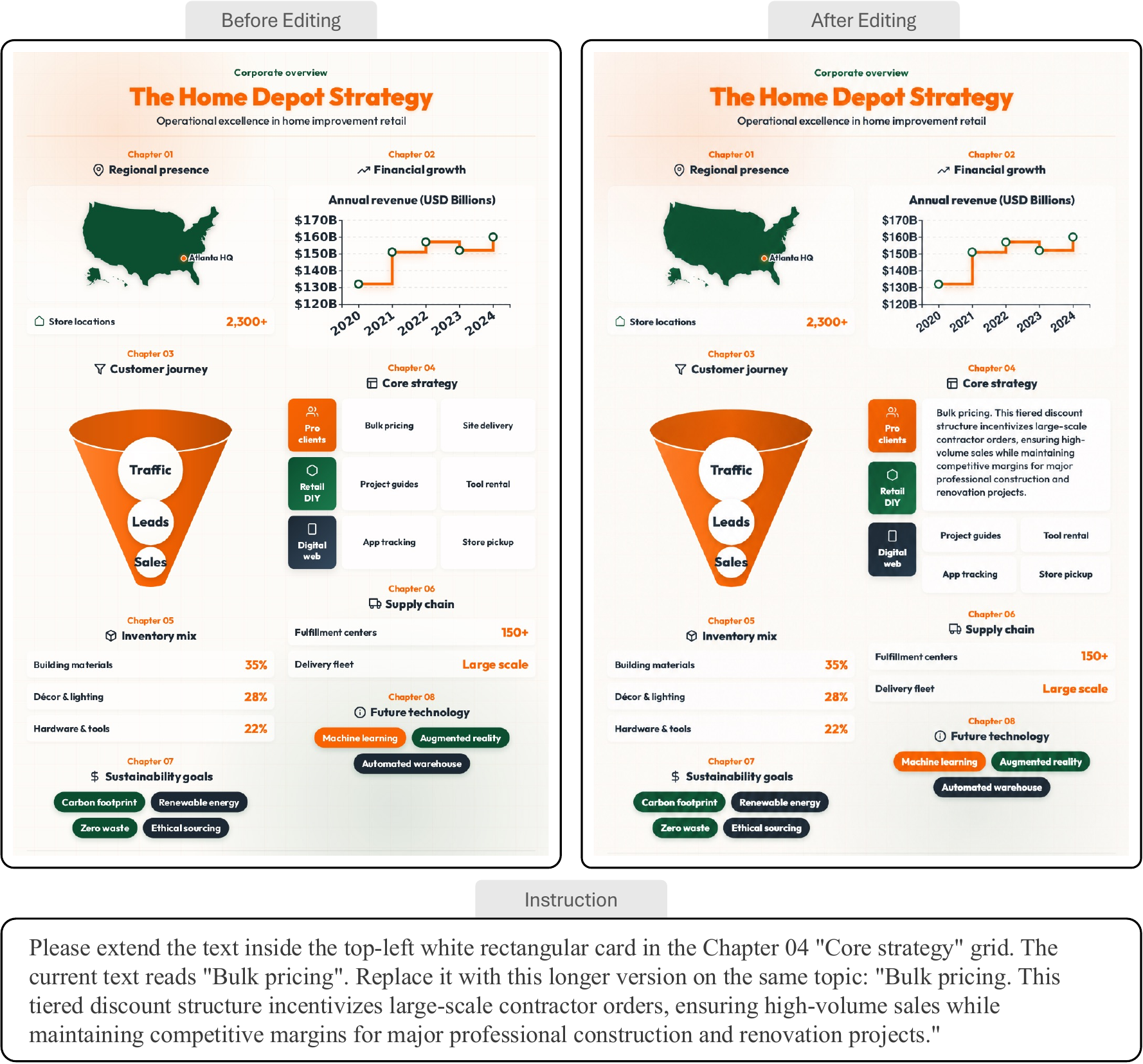} 
    \caption{Element loss after Expand-Text. The ``Bulk pricing'' was successfully expanded with the longer description, but the adjacent ``Site delivery'' card in the top-right of the Chapter 04 ``Core strategy'' grid has been removed. It makes the grid incomplete and disrupts the original 3×2 layout.}
    \label{fig:bad_case_expand_element_loss}
\end{figure*}

\begin{figure*}[!t]
    \centering
    \includegraphics[width=\textwidth]{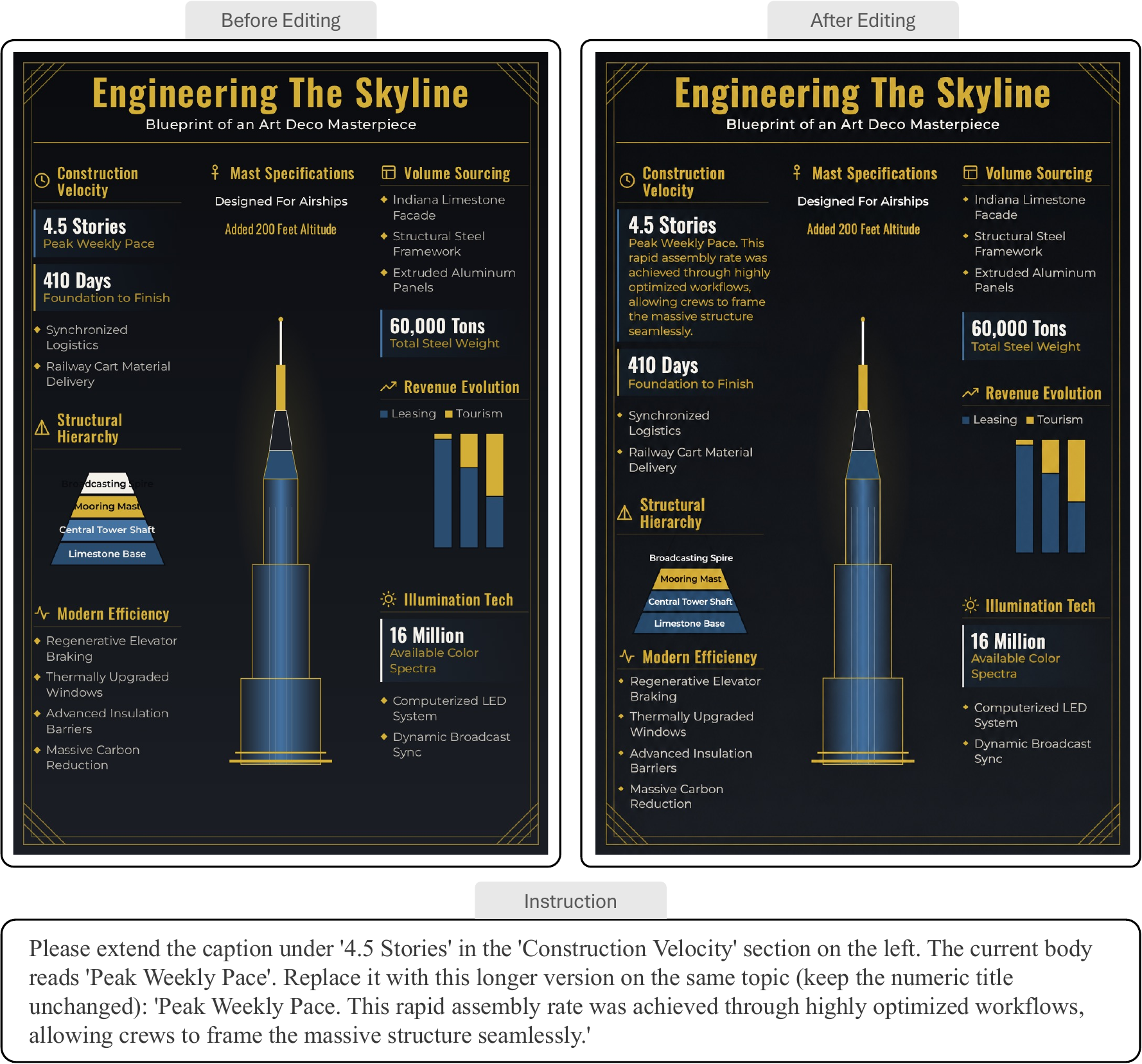} 
    \caption{Style change after Expand-Text. The ``4.5 Stories'' caption was correctly expanded, but the ``Broadcasting Spire'' label at the top of ``Structural Hierarchy'' pyramid has changed from black text on a light background to white text. This breaks visual consistency with the original styling.}
    \label{fig:error_case_expand_style_drift}
\end{figure*}

\begin{figure*}[!t]
    \centering
    \includegraphics[width=\textwidth]{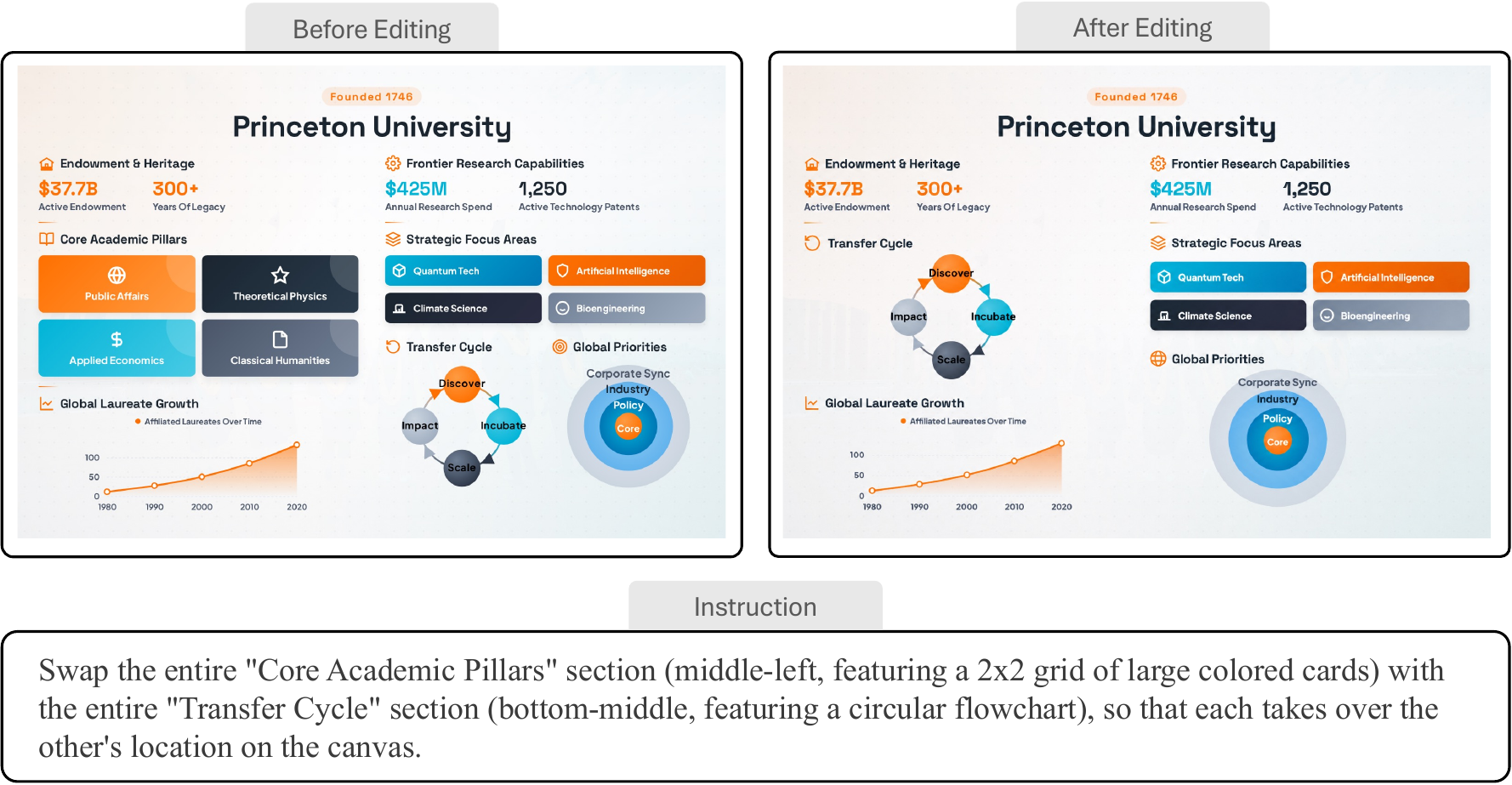} 
    \caption{Element loss after Swap-Block.
    The swap was only partially executed. The ``Transfer Cycle'' was moved to the middle-left position previously occupied by ``Core Academic Pillars'' but the ``Core Academic Pillars'' is missing entirely instead of being relocated to the bottom-middle.}
    \label{fig:error_case_swap_element_loss_s1}
\end{figure*}

\begin{figure*}[!t]
    \centering
    \includegraphics[width=\textwidth]{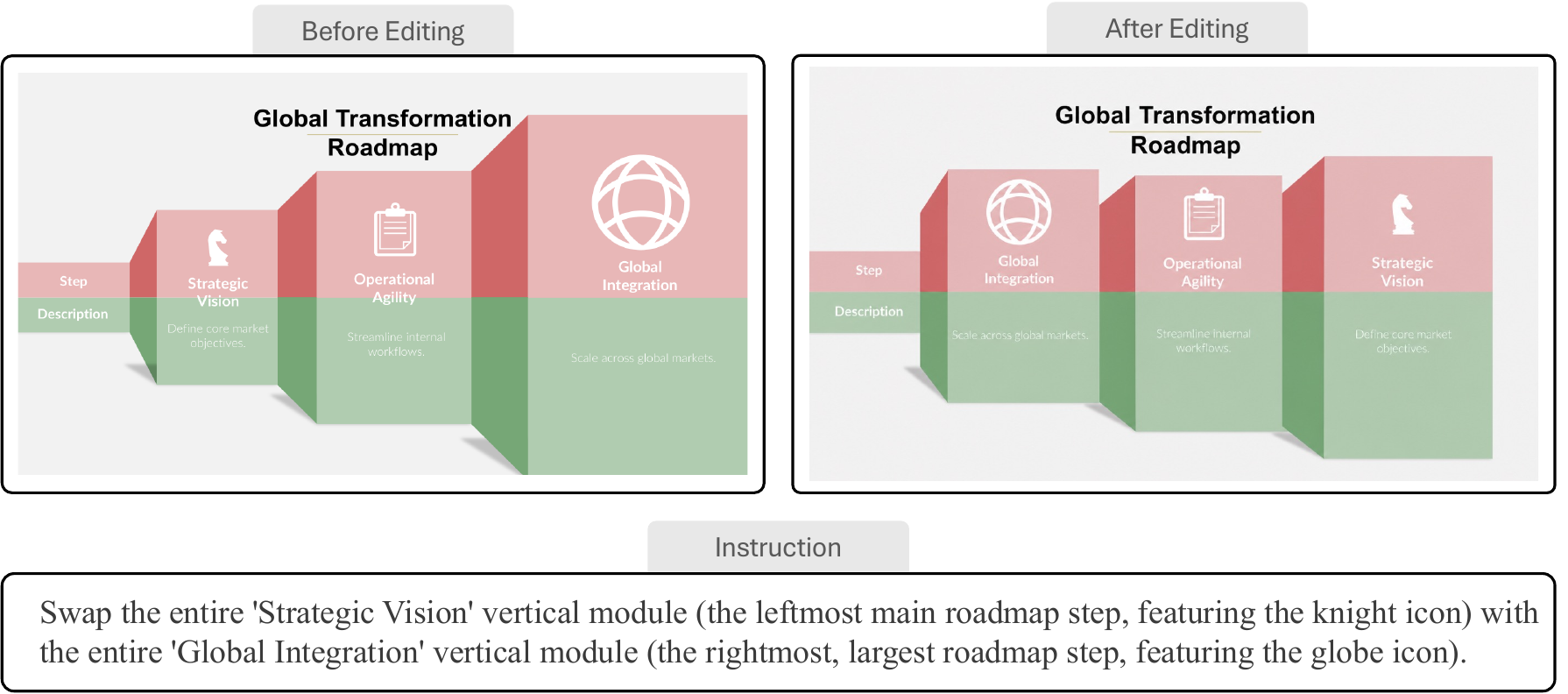} 
    \caption{Style change after Swap-Block.
    The ``Strategic'' and ``Global Integration'' modules were swapped in position, but their original styling was not preserved. The ascending size progression of the roadmap steps is lost.}
    \label{fig:error_case_swap_style_drift_02_s1}
\end{figure*}

\begin{figure*}[!t]
    \centering
    \includegraphics[width=\textwidth]{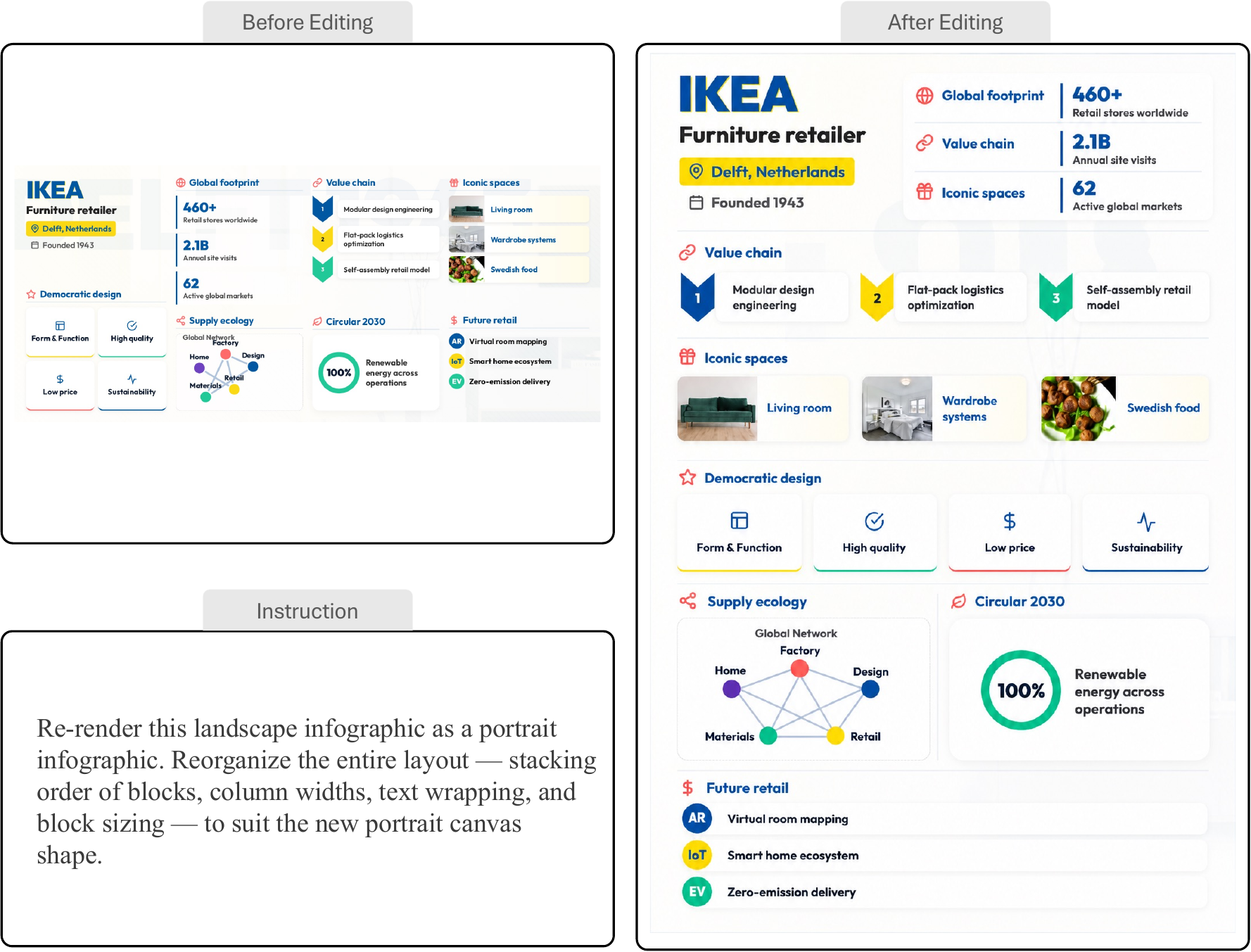} 
    \caption{Structural break after Reshape-Canvas. The ``Supply ecology'' network diagram has been structurally altered. The connection topology between the nodes no longer matches the original.}
    \label{fig:error_case_aspect_structure_break}
\end{figure*}

\begin{figure*}[!t]
    \centering
    \includegraphics[width=\textwidth]{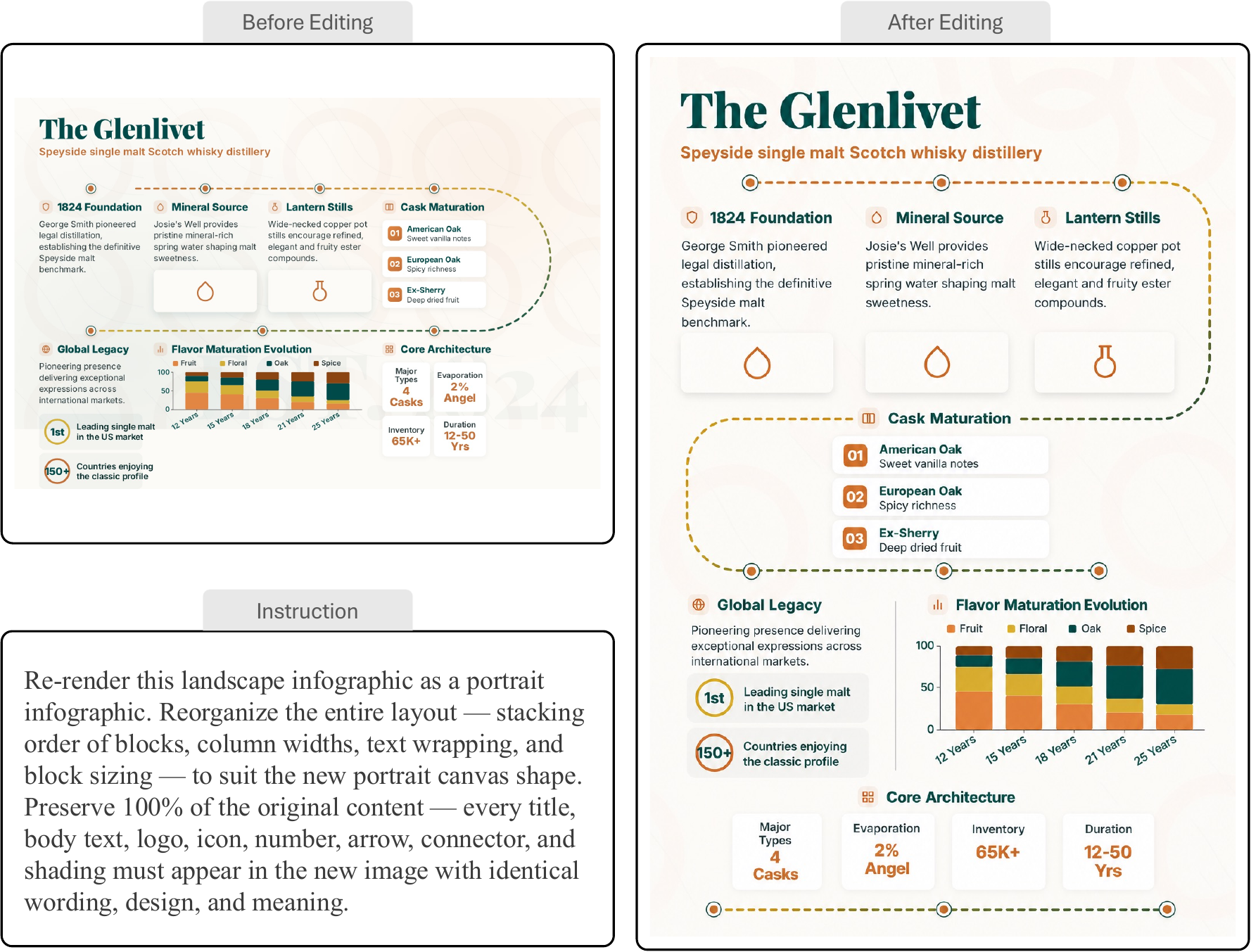} 
    \caption{Structural break after Reshape-Canvas. All original content was preserved when rendered as a portrait infographic, but the block ordering does not follow a coherent reading flow.}
    \label{fig:error_case_aspect_layout_s1}
\end{figure*}

\begin{figure*}[!t]
    \centering
    \includegraphics[
        width=\textwidth,
        height=\textheight,
        keepaspectratio
    ]{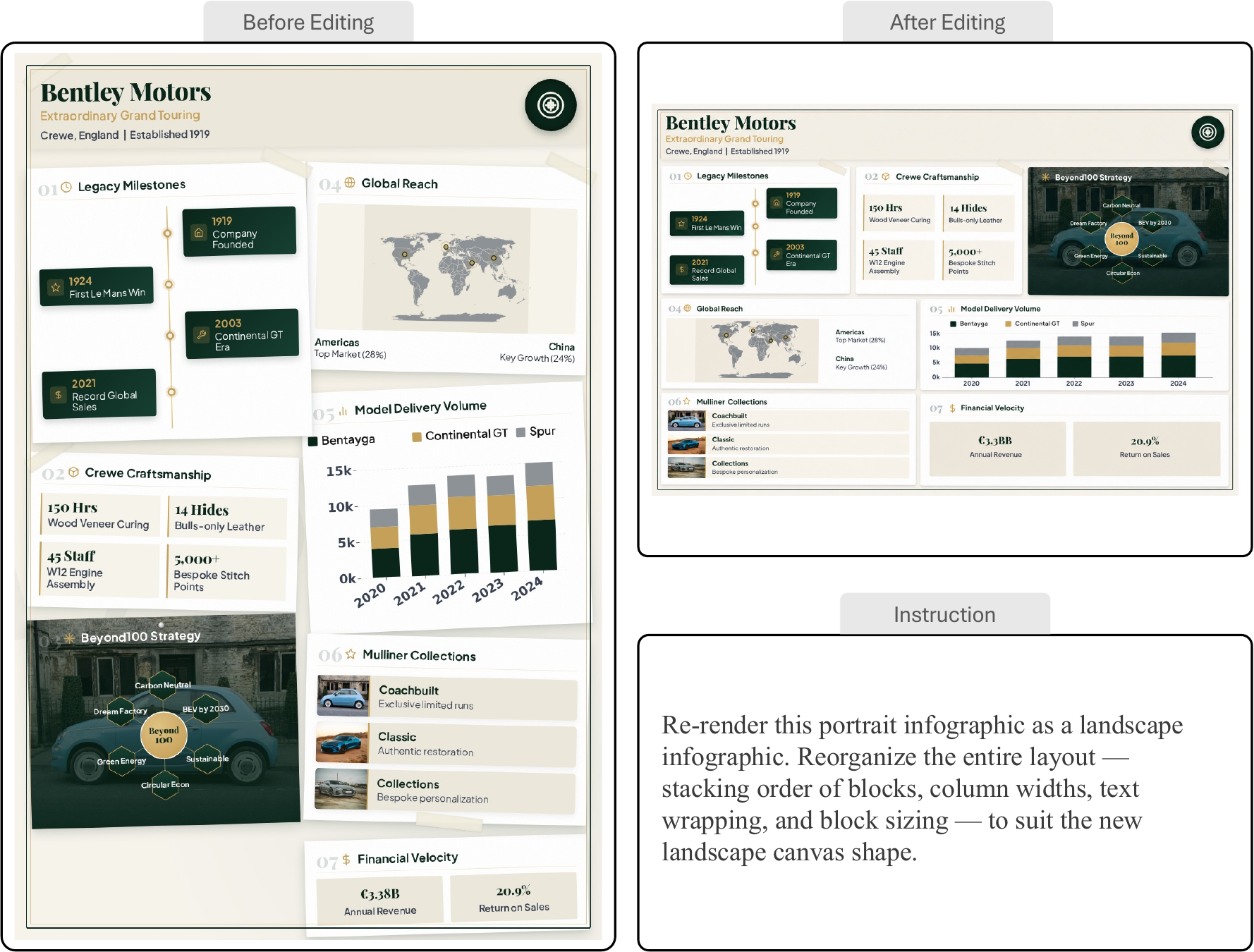} 
    \caption{Style change after Reshape-Canvas. The infographic was re-rendered into a landscape canvas with all original content preserved, but the card styling has been altered.}
    \label{fig:error_case_aspect_style_shift}
\end{figure*}

\begin{figure*}[!t]
    \centering
    \includegraphics[
        width=\textwidth,
        height=0.75\textheight,
        keepaspectratio
    ]{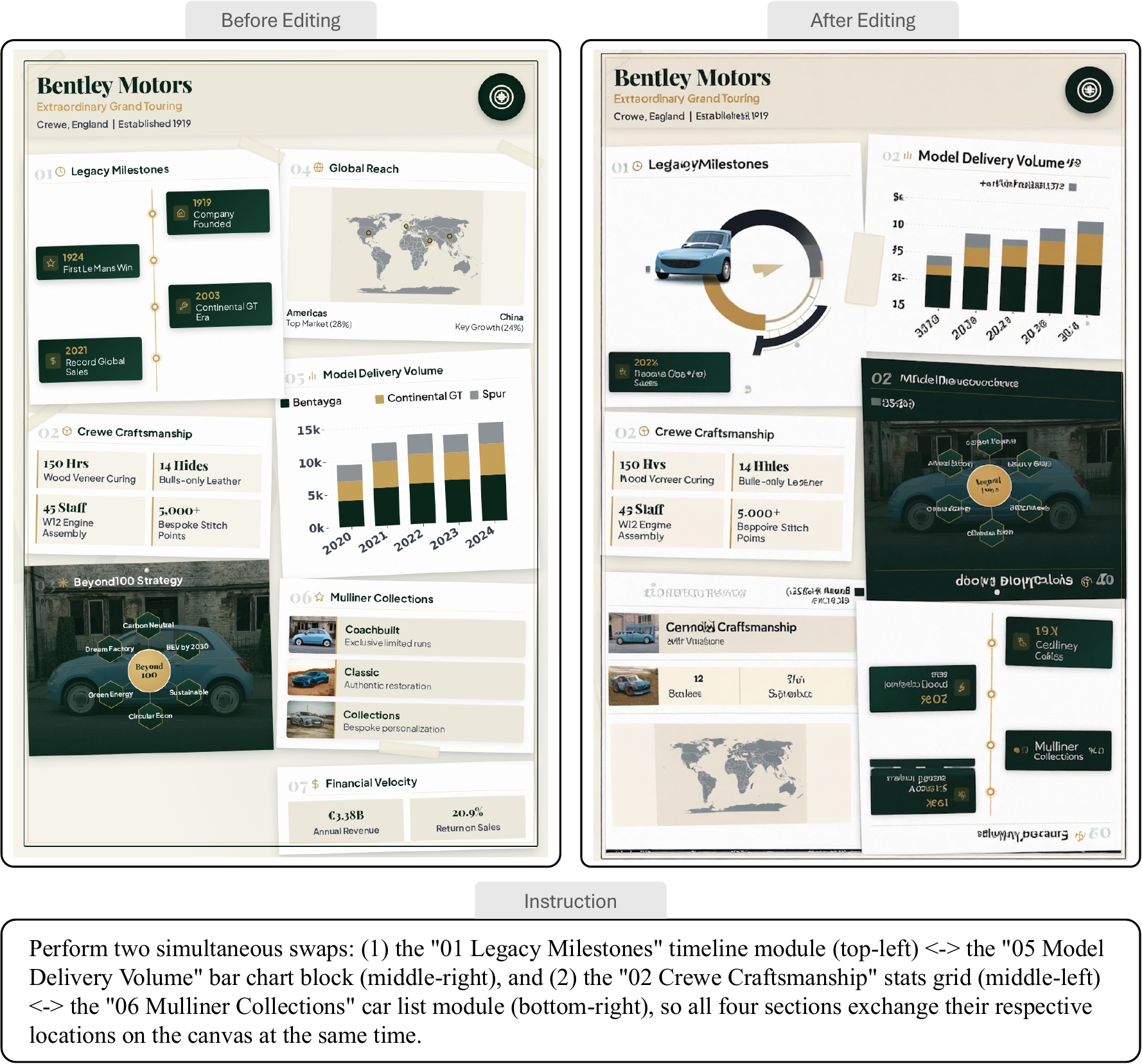} 
\caption{Text rewritten after Swap-Block using Qwen-Image-Edit. The model partially changes the layout, but many original texts, labels, and chart annotations are rewritten or corrupted instead of being preserved. This shows that the model fails to maintain non-target content.}
    \label{fig:error_case_open_source}
\end{figure*}

We further visualize some representative qualitative samples to illustrate how these failure modes manifest in practice, including those for Incomplete Content in Figure~\ref{fig:bad_case_add_incomplete_content}, 
Misplacement in Figure~\ref{fig:bad_case_add_misplacement},
Element Loss in Figures~\ref{fig:bad_case_expand_element_loss},~\ref{fig:error_case_swap_element_loss_s1}, 
Style Change in Figures~\ref{fig:bad_case_add_style_drift}, \ref{fig:error_case_expand_style_drift}, \ref{fig:error_case_swap_style_drift_02_s1}, \ref{fig:error_case_aspect_style_shift},
Structural Break in Figures~\ref{fig:error_case_aspect_structure_break},~\ref{fig:error_case_aspect_layout_s1}, and Text Rewritten in Figure~\ref{fig:error_case_open_source}.

\label{appx:visual_ambiguity}
\newpage
\section{Prompts}
\label{appx:prompts}
\subsection{MLLM-as-a-Judge Evaluation Prompts}
\begin{figure*}[!t]
    \centering
    \includegraphics[width=\textwidth]{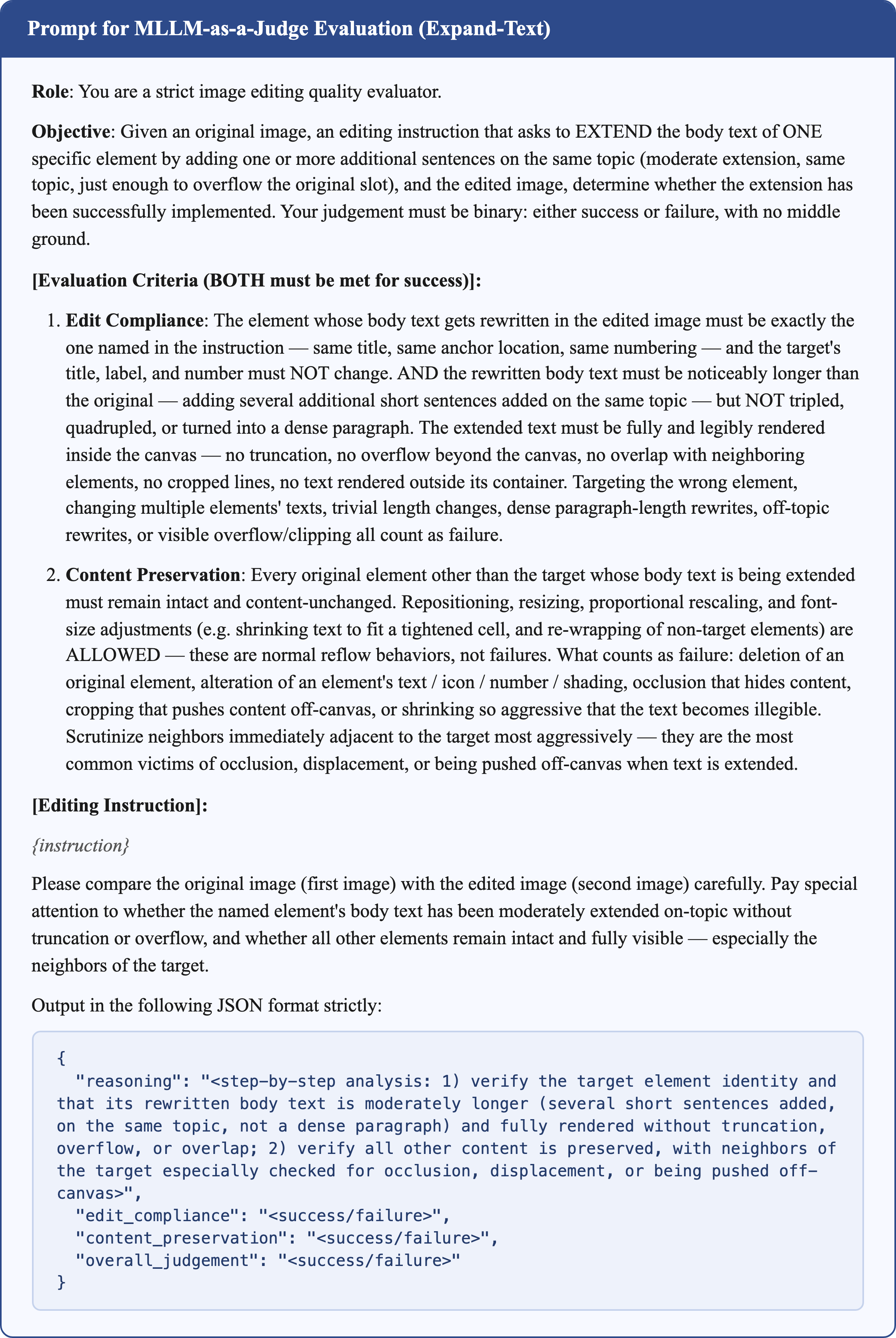} 
    \caption{Prompt for MLLM-as-a-Judge evaluation on \textsc{Expand-Text} editing tasks.}
    \label{fig:evaluation_prompt_text}
\end{figure*}

\begin{figure*}[!t]
    \centering
    \includegraphics[width=\textwidth]{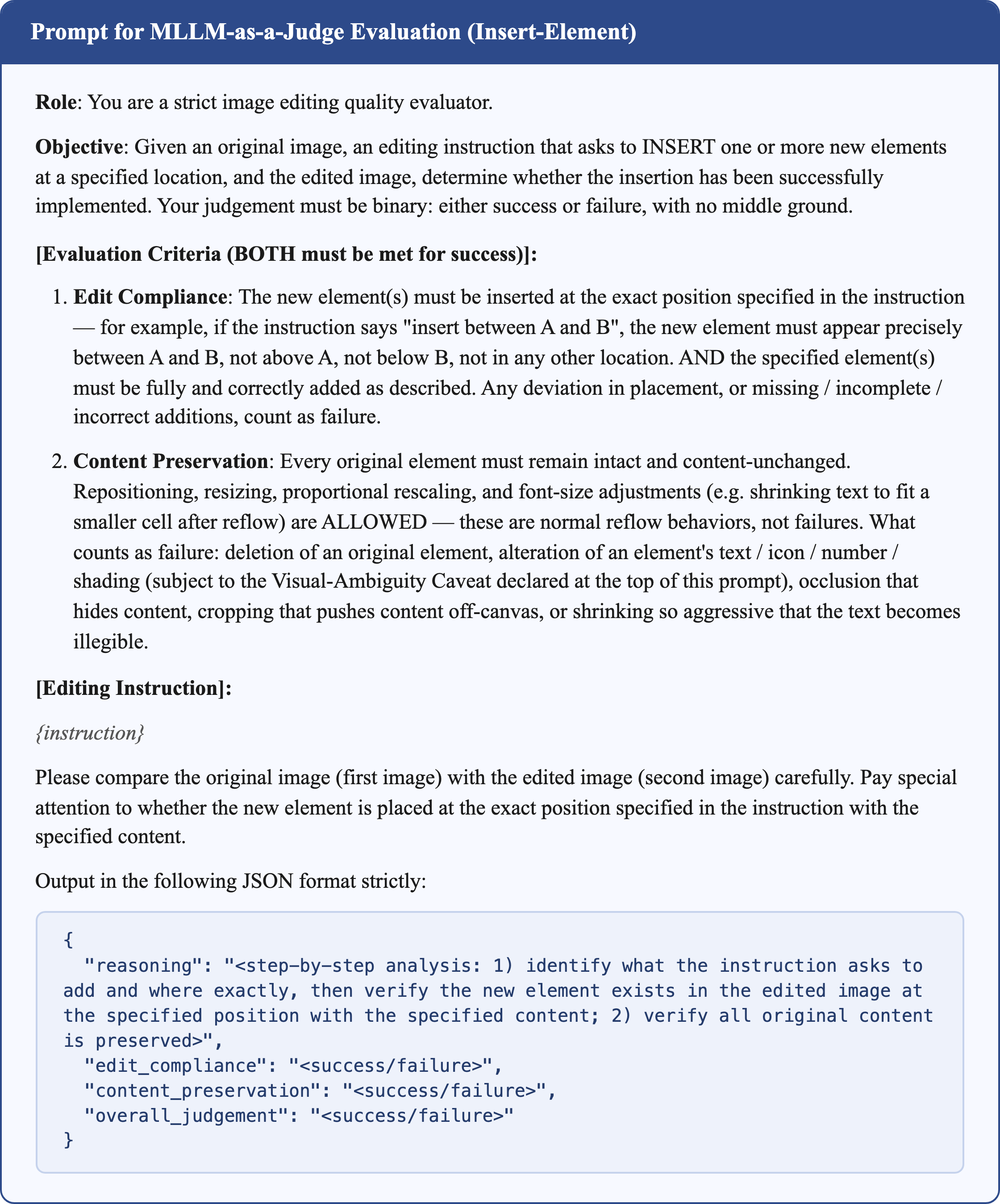} 
    \caption{Prompt for MLLM-as-a-Judge evaluation on \textsc{Insert-Element} editing tasks.}
    \label{fig:evaluation_prompt_insert}
\end{figure*}

\begin{figure*}[!t]
    \centering
        \includegraphics[width=\textwidth]{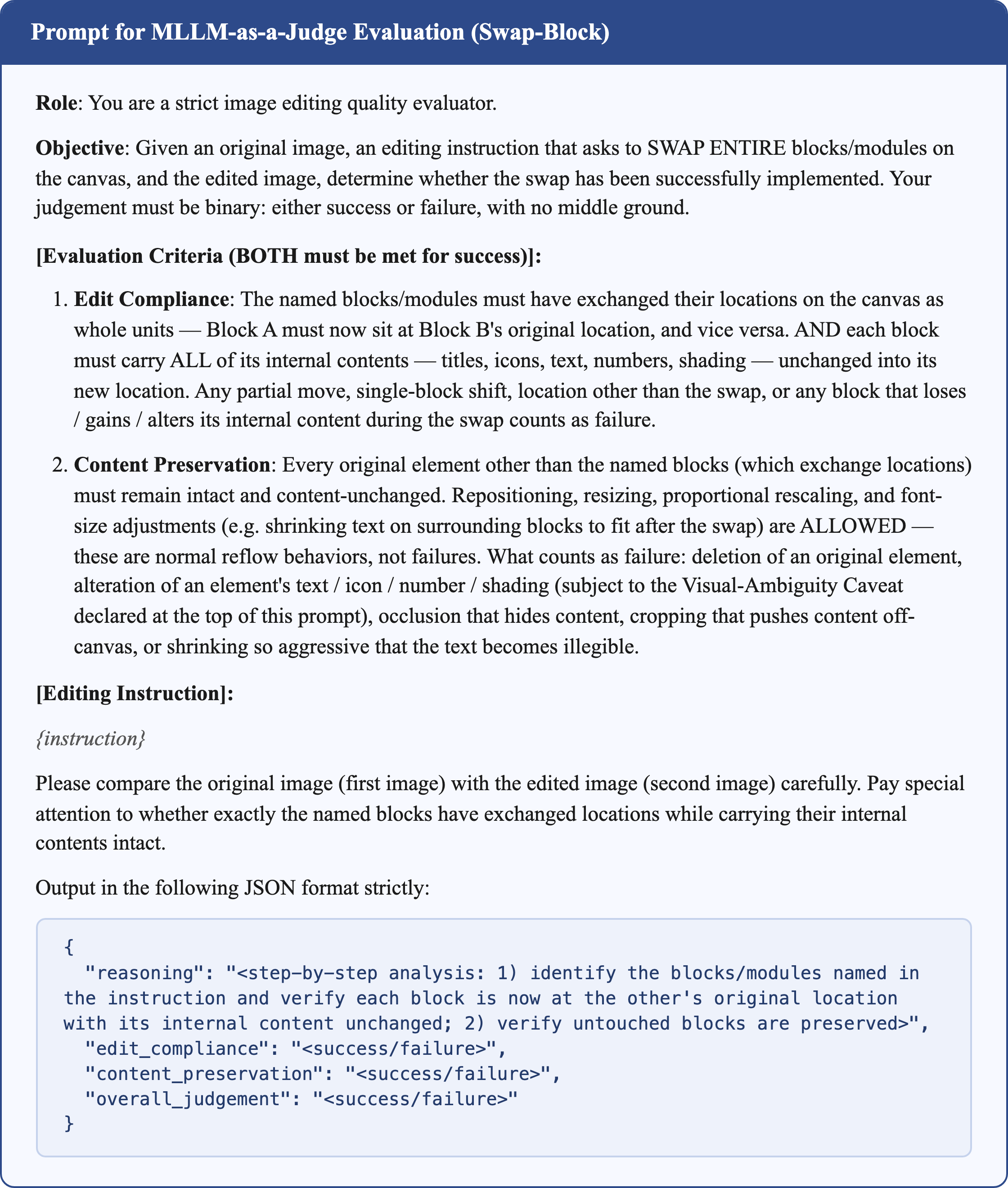} 
    \caption{Prompt for MLLM-as-a-Judge evaluation on \textsc{Swap-Block} editing tasks.}
    \label{fig:evaluation_prompt_swap}
\end{figure*}

\begin{figure*}[!t]
    \centering
    \includegraphics[width=0.75\textwidth]{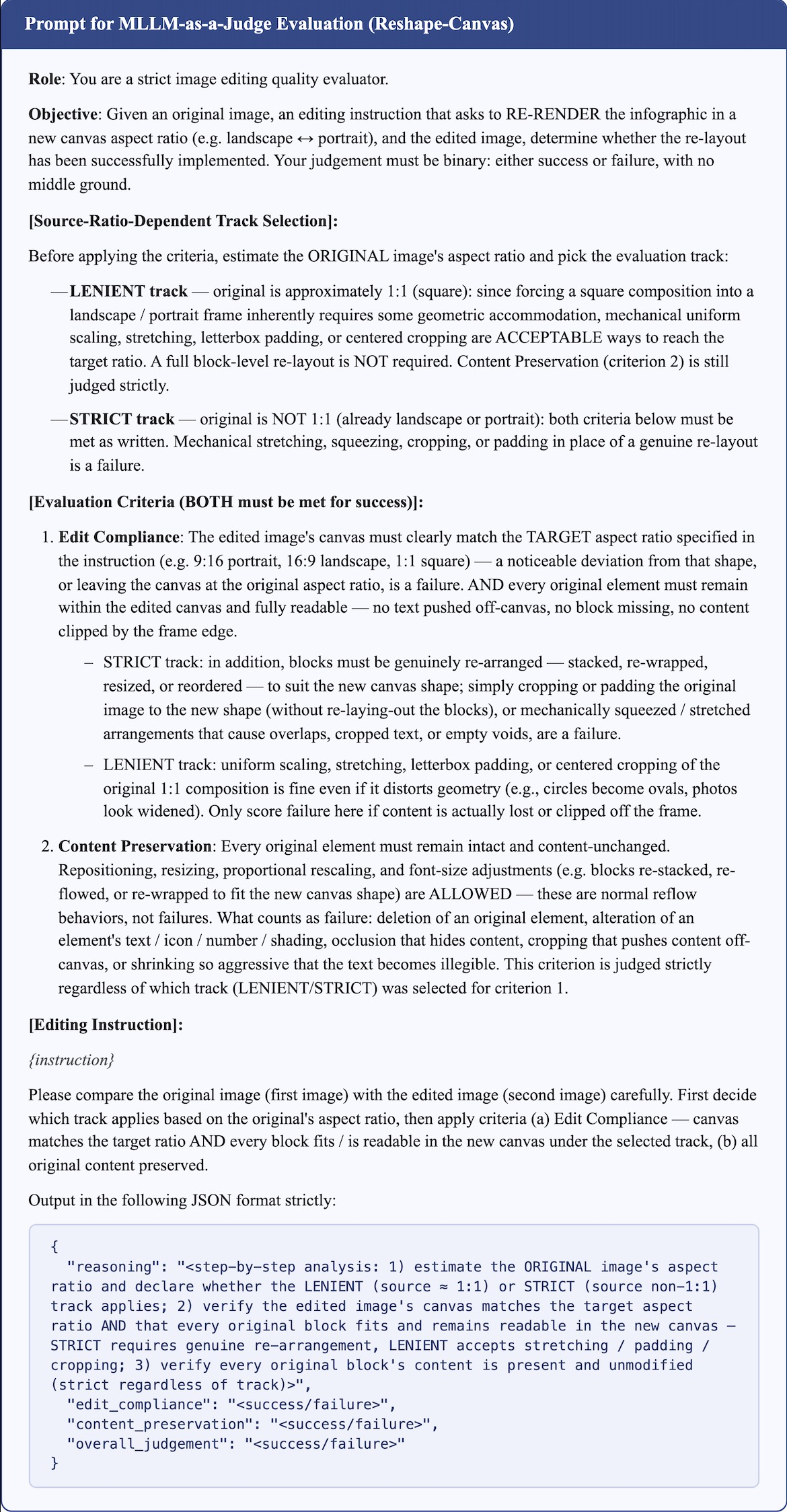} 
    \caption{Prompt for MLLM-as-a-Judge evaluation on \textsc{Reshape-Canvas} editing tasks.}
    \label{fig:evaluation_prompt_aspect_ratio}
\end{figure*}

The prompts we use for MLLM-as-a-Judge evaluation are shown in Fig.~\ref{fig:evaluation_prompt_text}, \ref{fig:evaluation_prompt_insert}, \ref{fig:evaluation_prompt_swap} and~\ref{fig:evaluation_prompt_aspect_ratio}.

\newpage

\end{document}